\documentclass[10pt, letter, onecolumn]{arxiv}

\usepackage{kantlipsum, lipsum}
\usepackage{dm-colors}
\usepackage{amsmath}
\usepackage{pstricks, pst-node}
\usepackage{verbatim}
\usepackage{multirow}
\usepackage{array}
\usepackage{longtable}
\usepackage{pdflscape}
\usepackage{scalerel}
\usepackage{booktabs}
\usepackage{enumitem}
\usepackage{xspace}
\usepackage{bm}
\usepackage{bbm}
\usepackage{mathtools}
\usepackage{soul}
\usepackage{epsfig}
\usepackage{graphicx}
\usepackage{amssymb}
\usepackage{colortbl}
\usepackage{csquotes}
\usepackage{setspace}
\usepackage{colortbl}
\usepackage{bbding}
\usepackage{threeparttable}
\usepackage{tabularx,ragged2e}
\usepackage{placeins}
\usepackage{bbding}

\definecolor{darkpastelgreen}{rgb}{0.13, 0.55, 0.13}
\definecolor{darkpastelred}{rgb}{0.55, 0.13, 0.13}
\usepackage[hang,flushmargin]{footmisc}

\usepackage{nameref}
\usepackage{varioref}
\usepackage{amssymb}%
\usepackage{lineno}
\usepackage{graphicx}
\usepackage{algorithm}
\usepackage{algpseudocode}

\usepackage{pifont}

\usepackage[pagebackref=false,breaklinks=false,%
            colorlinks=true,bookmarks=true,citecolor=ourdarkblue,%
            urlcolor=ourdarkblue,linkcolor=ourdarkblue]{hyperref}
\usepackage[noabbrev,capitalize]{cleveref}
\usepackage{etoc}
\usepackage{lineno}
\usepackage{algorithm}
\usepackage{makecell} 
\usepackage{algpseudocode}

\usepackage{amsmath}
\usepackage{thmtools}
\usepackage[most]{tcolorbox}
\usepackage{lmodern}

\usepackage[most]{tcolorbox}
\usepackage{lmodern}

\declaretheoremstyle[
    spaceabove=6pt, spacebelow=6pt,
    headfont=\bfseries, headpunct={.}, headformat={\NAME\ \NUMBER},
    bodyfont=\normalfont,
    postheadspace=0.5em
]{promptstyle}

\declaretheorem[name=Prompt, style=promptstyle]{prompt}

\declaretheorem[name=Case, style=promptstyle]{case}

\tcolorboxenvironment{prompt}{
    colback=softblue!5!white, 
    colframe=softblue!40!white, 
    fonttitle=\bfseries,
    title=Prompt,
    boxrule=0.5pt,
    sharp corners,
    breakable
}

\tcolorboxenvironment{case}{
    colback=softblue!5!white, 
    colframe=softblue!40!white, 
    fonttitle=\bfseries,
    title=Case,
    boxrule=0.5pt,
    sharp corners,
    breakable
}

\graphicspath{{figures/}}
\definecolor{midgreen}{rgb}{0.0, 0.75, 0.0}

\definecolor{goodgreen}{HTML}{00B050}
\definecolor{badred}{HTML}{FF0000}
\definecolor{softblue}{HTML}{136783}

\newcommand{\benchname}{EHR-Bench\xspace}
\newcommand{\modelname}{EHR-R1\xspace}
\newcommand{\traindataname}{EHR-Ins\xspace}

\title{\Large{EHR-R1: A Reasoning-Enhanced Foundational Language Model for Electronic Health Record Analysis}}

\author[$\ast$,1]{Yusheng Liao} 
\author[$\ast$,1]{Chaoyi Wu} 
\author[$\ast$,3,4]{Junwei Liu} 
\author[2,5]{Shuyang Jiang} 
\author[1,2]{Pengcheng Qiu}
\author[3]{\\Haowen Wang}
\author[3]{Yun Yue}
\author[3]{Shuai Zhen}
\author[3]{Jian Wang}
\author[3]{Qianrui Fan}
\author[3]{Jinjie Gu}
\author[1,2]{\\ \vspace{0.1cm} Ya Zhang} 
\author[1,2]{Yanfeng Wang} 
\author[1,2,$\dag$]{Yu Wang} 
\author[1,2,$\dag$]{Weidi Xie}
\affil[1]{\normalsize Shanghai Jiao Tong University, Shanghai, China \authorcr \vspace{0.1cm}}

\affil[2]{\normalsize Shanghai Artificial Intelligence Laboratory, Shanghai, China  \authorcr \vspace{0.1cm}}

\affil[3]{\normalsize Intelligence Healthcare Department, AntGroup, Hangzhou, China  \authorcr \vspace{0.1cm}}

\affil[4]{\normalsize Intelligence Computing and Sensing Laboratory, Peking University, Beijing, China  \authorcr \vspace{0.1cm}}

\affil[5]{\normalsize Fudan University, Shanghai, China \authorcr \vspace{0.1cm}}

\affil[$\ast$]{\normalsize Equal contributions\hspace{1cm}}
\affil[$\dag$]{\normalsize Corresponding author\authorcr Yu Wang: yuwangsjtu@sjtu.edu.cn; Weidi Xie: weidi@sjtu.edu.cn}

\begin{document}

\begin{abstract}
Electronic Health Records (EHRs) contain rich yet complex information, and their automated analysis is critical for clinical decision-making. Despite recent advances of large language models (LLMs) in clinical workflows, their ability to analyze EHRs remains limited due to narrow task coverage and lack of EHR-oriented reasoning capabilities.
This paper aims to bridge the gap, specifically, we present \textbf{EHR-Ins}, a large-scale, comprehensive EHR reasoning instruction dataset, comprising 300k high-quality reasoning cases and 4M non-reasoning cases across 42 distinct EHR tasks. 
Its core innovation is a \textbf{thinking-graph-driven framework that enables to generate high-quality reasoning data} at scale.
Based on it, we develop \textbf{EHR-R1}, a series of reasoning-enhanced LLMs with up to 72B parameters tailored for EHR analysis. Through a multi-stage training paradigm, including \textbf{domain adaptation}, \textbf{reasoning enhancement}, and \textbf{reinforcement learning}, EHR-R1 systematically acquires domain knowledge and diverse reasoning capabilities, enabling accurate and robust EHR analysis.
Lastly, we introduce \textbf{\benchname}, a new benchmark curated from MIMIC-IV, spanning 42 tasks, to comprehensively assess reasoning and prediction across EHR scenarios. 
In experiments, we show that the resulting EHR-R1 consistently outperforms state-of-the-art commercial and open-source LLMs (including DeepSeek-V3 and GPT-4o), surpassing GPT-4o by over 30 points on EHR-Bench and achieving a 10\% higher zero-shot AUROC on EHRSHOT. Collectively, EHR-Ins, EHR-R1, and EHR-Bench have significantly advanced the development for more reliable and clinically relevant EHR analysis.

\end{abstract}

\maketitle


\section{INTRODUCTION}

Electronic Health Records (EHRs) are comprehensive digital repositories of patient information, encompassing laboratory tests, medications, diagnoses, procedures, clinical notes, \emph{etc.}~\cite{sheikhalishahi2020benchmarking,DBLP:conf/nips/YecheKZHLFR21,DBLP:journals/jbi/SolaresRZRCTGPZ20,DBLP:conf/amia/XieZLTLRCCWDOGL22,DBLP:conf/chil/WangMCGHN20}. Systematic analysis of EHRs is essential for modern healthcare, as accurate interpretation of patient histories enables early disease detection, personalized treatment planning, and improved clinical outcomes. In everyday clinical practice, physicians rely on electronic health records (EHRs) to address a wide spectrum of analytical needs, spanning decision-support queries, for example, \emph{``what is the most probable next diagnosis for this patient?''}, to prognostic assessments, such as \emph{``what is the risk of hospital readmission within 60 days?''} Advanced EHR analysis systems can substantially accelerate clinical workflows and augment medical decision-making, ultimately shaping the quality and timeliness of patient care.

Large language models (LLMs) have recently transformed biomedical natural language processing~(BioNLP), achieving impressive results across a range of medical tasks~\cite{nori2023capabilities, singhal2023large, liao2024automatic, qiu2025quantifying, DBLP:journals/npjdm/WuQLGLZWX25, DBLP:conf/acl/LiaoJW025, bedi2025medhelm, wu2025medreason}, yet their performance on EHR-related tasks remains limited~\cite{rouhizadeh2025large, qu2024enhancing, soroush2024large, ren2025comprehensive}. Even state-of-the-art commercial models struggle to extract, integrate, and reason over EHR data, which markedly limits their interoperability with daily used hospital information systems~\cite{zhu2025healthflow} and, consequently, their wide adoption in clinical practice.

Existing work on EHR analysis with LLMs is fundamentally constrained by two major challenges. First, in terms of \textbf{task coverage}, prior research has focused on narrow, task-specific objectives (\emph{e.g.}, risk prediction for a specific condition or outcome), typically restricted to particular diseases or event types, and has yet to deliver the holistic capabilities required to support evolving clinical workflows~\cite{NEURIPS2023_d42db1f7,lin2025training,hegselmann2025large}. Second, in terms of \textbf{reasoning ability}, existing models struggle to construct reliable, EHR-oriented reasoning chains that demand both selective information filtering and multi-source integration. As a result, they are highly susceptible to the redundancy inherent in EHR data~\cite{wu2024instruction,chen2025narrative} and struggle to integrate isolated findings into a coherent longitudinal understanding of disease progression~\cite{wu2024instruction,kim2023detection}.

\begin{figure}[!t]
    \centering
    \includegraphics[width=1.0\linewidth]{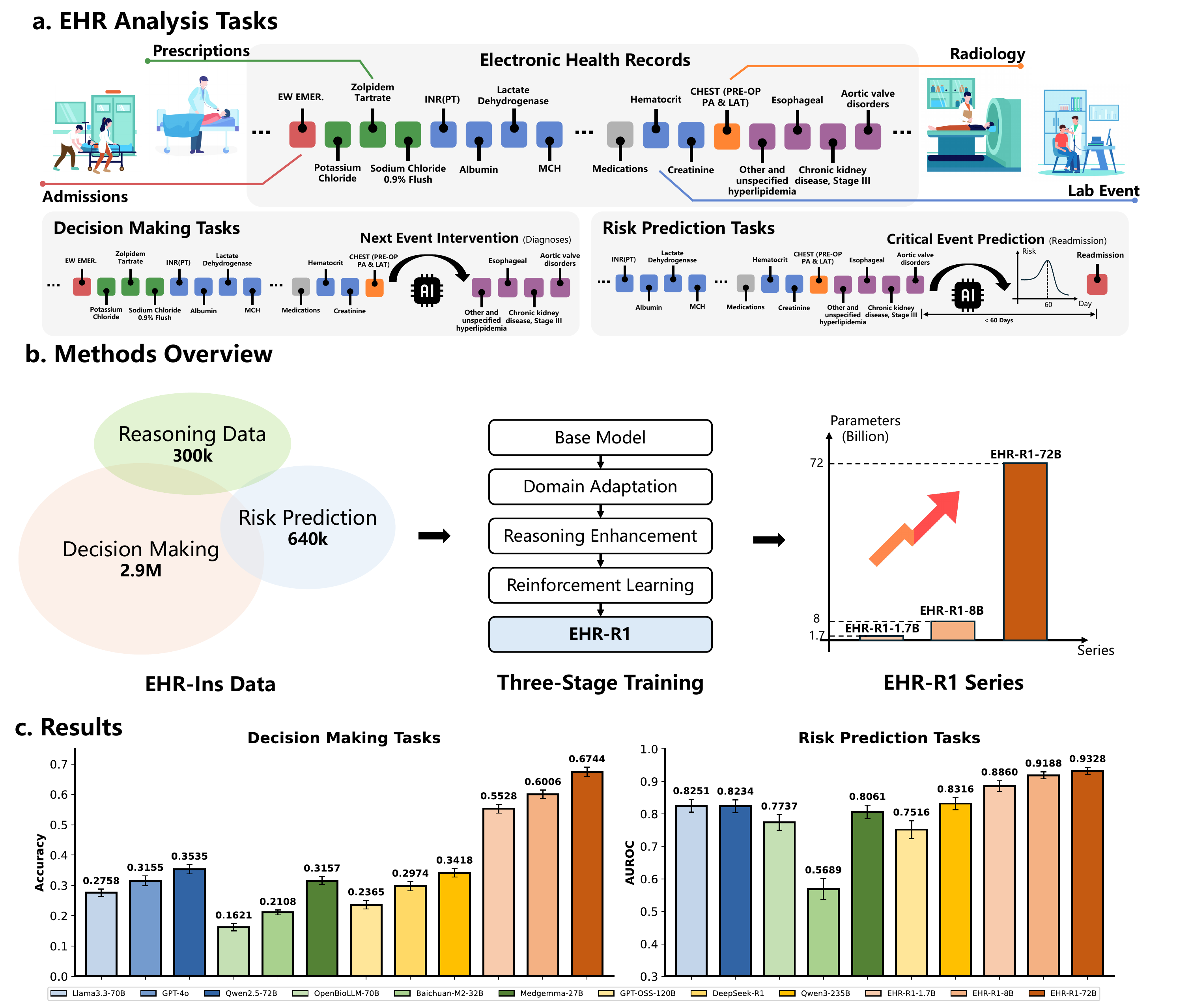}
    \caption{\textbf{Overview of the EHR tasks and the proposed method.} \textbf{a. EHR Analysis Tasks.} EHR analysis tasks are defined as consisting of two types of tasks: decision-making and risk-prediction. \textbf{b. Methods Overview.} Our approach addresses these challenges with a three-stage training pipeline. First, a large volume of non-reasoning data is used for continual pre-training. This is followed by an instruction-tuning phase that leverages reasoning data. Finally, reinforcement learning with Group Reward Policy Optimization (GRPO) is applied to further refine the model. \textbf{c. Results.} This figure compares the performance of our model against several baseline LLMs on both decision-making and risk-prediction tasks, showcasing its superior performance.}
    \label{fig: teaser}
\end{figure} 

In this paper, to address these, we propose a holistic framework for EHR analysis with two primary contributions: 
(i) a large-scale super-instruction dataset that unifies diverse EHR analysis tasks into a generative format and incorporates reasoning supervision, and (ii) a reasoning-enhanced EHR analysis LLM trained with synthetic reasoning data via a three-stage training paradigm. 
The overview of the EHR tasks and our proposed method are shown in Figure~\ref{fig: teaser}.

We present \textbf{\traindataname}, a large-scale dataset capturing broad task diversity and explicit medical reasoning for EHR analysis. The corpus comprises \textbf{300K} high-quality reasoning cases and \textbf{3.5 million} non-reasoning cases spanning 42 EHR tasks. 
The tasks generally fall into two categories: 
\textbf{decision-making}~({\em e.g.}, diagnosis and treatment recommendations) and \textbf{risk-prediction}~({\em e.g.}, mortality and readmission). To ensure clinical fidelity and relevance, we develop a `thinking-graph' driven reasoning data synthesis pipeline that (i) applies statistical analysis of entity co-occurrence ratios to identify key related entities, (ii) links them via knowledge from UMLS~\cite{bodenreider2004unified}, and (iii) prompts GPT-4o to produce structured, step-by-step clinical reasoning. The resulting dataset is extensive and clinically grounded, enabling models to acquire diverse, context-rich reasoning capabilities.

Second, we introduce \textbf{\modelname}, a family of reasoning-enhanced LLMs (up to \textbf{72B} parameters) tailored for EHR analysis and trained on \traindataname. The training curriculum is designed to enhance domain knowledge and diverse reasoning patterns through three stages: 
(i) large-scale domain adaptation on extensive non-reasoning data, 
(ii) reasoning enhancement on high-quality reasoning cases, 
and (iii) reinforcement learning with Group Relative Policy Optimization (GRPO) on a smaller, curated set. 
This multi-stage regime improves the models’ ability to handle complex EHR tasks and to produce accurate, clinically meaningful outputs across diverse scenarios.

We evaluate \textbf{\modelname} on EHRSHOT~\cite{NEURIPS2023_d42db1f7} and MIMIC-IV-CDM~\cite{hager_evaluation_2024}, as well as on a new benchmark, \textbf{\benchname}, derived from MIMIC-IV. \benchname spans 42 tasks across decision-making and risk-prediction settings, providing a balanced, comprehensive assessment of both reasoning and task-specific performance. Together, these benchmarks cover two distinct clinical centers—Stanford Medicine and Beth Israel Deaconess Medical Center—and encompass a broad range of EHR analysis tasks.

Experimental results demonstrate that our model consistently outperforms leading commercial and open-sourced LLMs. 
In particular, \modelname-72B achieves an average performance improvement of over \textbf{30 points} compared to GPT-4o across all 42 tasks on \benchname, highlighting the effectiveness of our method in enhancing LLMs to tackle the full spectrum of EHR-related challenges. On the out-of-distribution EHRSHOT dataset, which features significantly different structures and medical concepts, \modelname-72B achieves a zero-shot AUROC score that is 10\% higher than baseline models. These results highlight not only the superior task adaptability of \modelname, but also its robustness and generalizability to unseen datasets, underscoring its potential as a transformative tool for clinical decision support.

\section{Results}

In the following, we first give an overview on the data quality of our constructed training data, \traindataname, then describe the benchmarks and evaluation metrics. Afterwards, we report the performance of our final reasoning-enhanced model, \modelname, highlighting its consistent gains over leading LLMs.

\begin{figure}[!t]
    \centering
    \includegraphics[width=1.\linewidth]{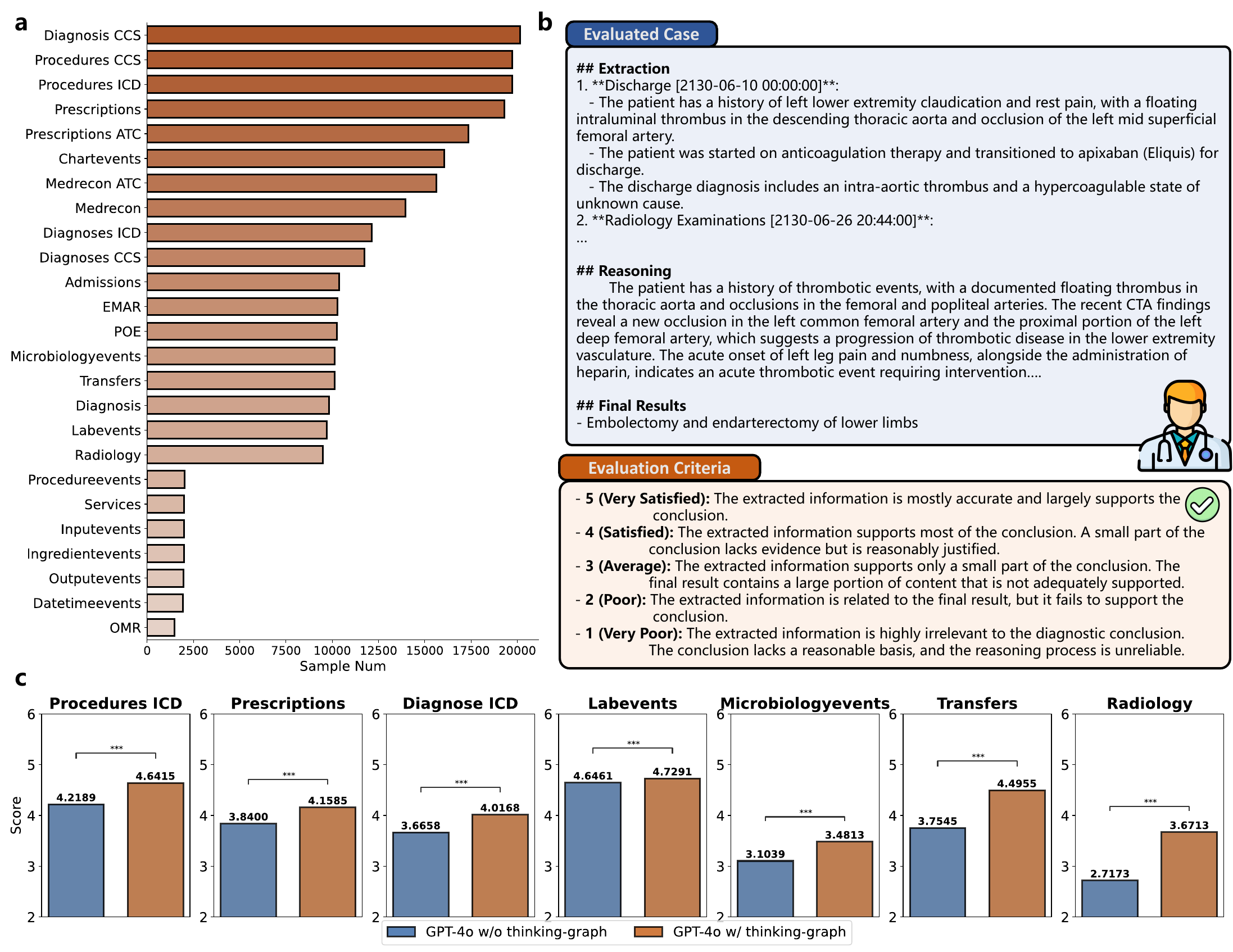}
    \caption{\textbf{Overview of reasoning data in \traindataname.} \textbf{a} The sample size of each task in the reasoning data of \traindataname. \textbf{b} Example of human evaluation on the reasoning data. \textbf{c} Manual evaluation results on EHR reasoning data across eight decision-making tasks, each associated with a distinct type of decision-making event. We compared the quality of synthetic reasoning data with and without thinking-graph enhancement, where `***' represents a significance level of $p<0.001$.}
    \label{fig: reasoning}
\end{figure}

\subsection{Training Data Overview}
To enhance the LLM's understanding of EHR data and its ability to identify relationships between medical entities within redundant EHR records, we propose a novel `thinking-graph' pipeline to curate a new large-scale reasoning-enhanced EHR-analysis instruction data, \textbf{\traindataname}. This pipeline is designed to augment GPT-4o~\cite{hurst2024gpt}, enabling it to generate high-quality EHR reasoning chains. The details of this methodology can be found in Section~\ref{section: reasoning synthesis}. The synthesized reasoning data distribution for each analysis task in \traindataname is illustrated in Figure~\ref{fig: reasoning}a, widely covering diverse decision-making tasks.

To demonstrate the effectiveness of our thinking‑graph pipeline in synthesizing reasoning chains for EHR analysis, we hired medical experts to manually, evaluate 100 synthesized reasoning chains across eight decision-making tasks to check the data quality. As shown in Figure~\ref{fig: reasoning}b, the experts are tasked to rate each generated reasoning samples on a 5-point scale. The scores from highest to lowest represent how completely the reasoning process supports the predicted label set. A score of 5 indicates that the reasoning process fully supports the predicted result, while a score of 1 indicates the reasoning process is completely irrelevant to the predicted result. Alongside our \emph{thinking‑graph} pipeline, we adopt a naïve data distillation strategy with GPT‑4o as a baseline, \emph{i.e.}, directly prompting it (using Prompt~\ref{prompt: evaluation}) to derive reasoning chains from the original EHR analysis cases. 

The corresponding results are shown in Figure~\ref{fig: reasoning}c. First, \textbf{in relative comparison}, the \emph{thinking‑graph} pipeline produced reasoning chains with more adequate EHR evidence across all eight tasks than the naïve data distillation approach. These findings demonstrate that our extracted \emph{thinking‑graph} yields more reliable references for GPT‑4o, effectively mitigating its knowledge limitations and enhancing curated data quality.
Second, \textbf{in terms of absolute scores}, the medical experts expressed satisfaction with most of the synthesized reasoning data, assigning it an average rating above 4 on a five‑point scale. This evaluation underscores the ability of the thinking‑graph pipeline to effectively extract and utilize sufficient medical evidence from EHR data to support target entities, thereby substantially enhancing the capabilities of the LLM.


\begin{figure}[!t]
    \centering
    \includegraphics[width=1.\linewidth]{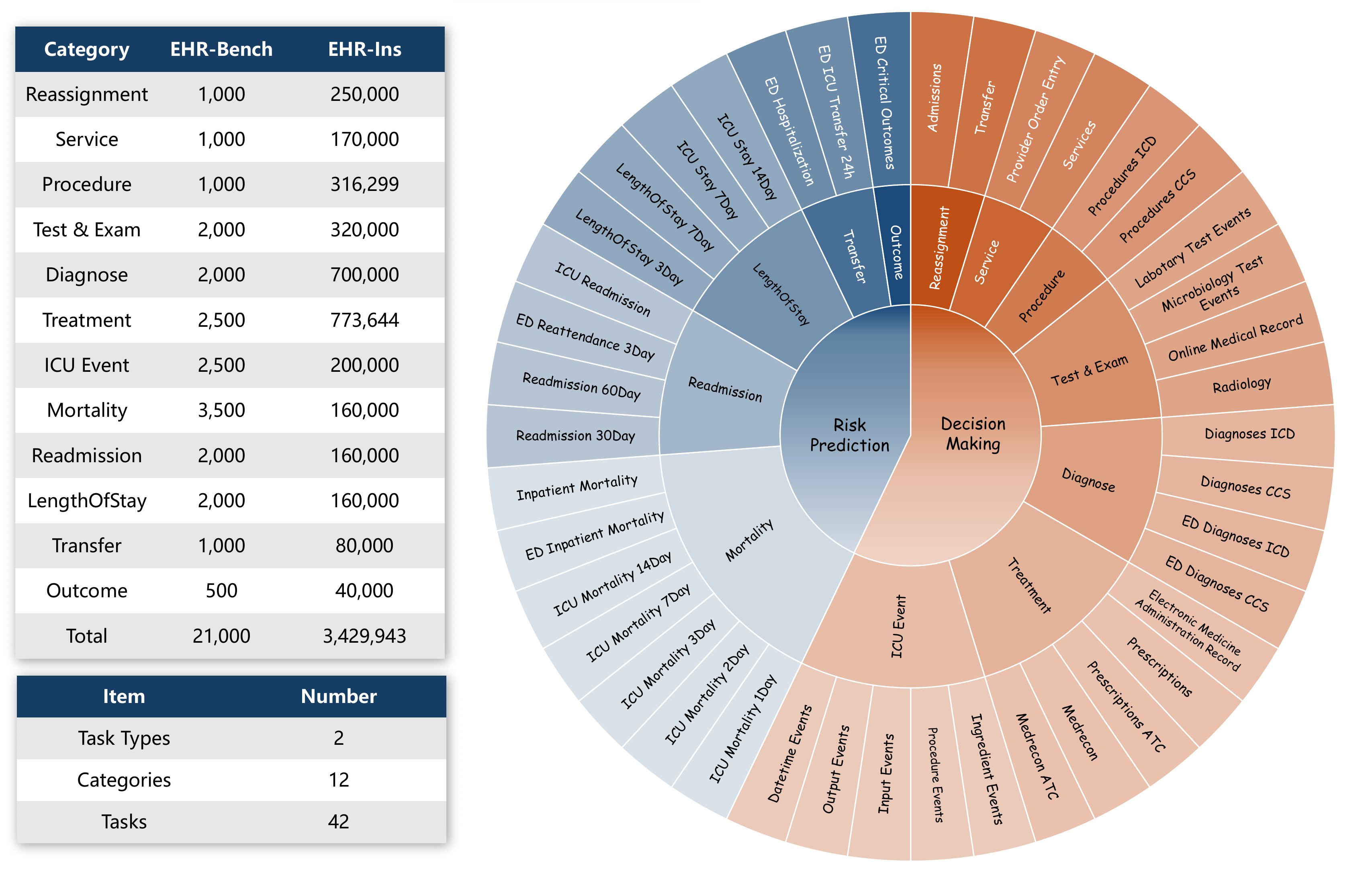}
    \caption{\textbf{Overview of \traindataname and  \benchname.}. The hierarchical ring chart displays the distribution of both datasets. The inner ring partitions tasks into two types: risk prediction and decision making. The middle ring shows 12 task categories (subtypes). The outer ring details all 42 specific tasks.}
    \label{fig: data overview}
\end{figure} 

\subsection{Evaluation Setting}
\subsubsection{Benchmarks}

We evaluate on three benchmarks to comprehensively assess capabilities. \benchname is our primary in-distribution benchmark, constructed from the same MIMIC-IV source as training and providing a holistic assessment across diverse EHR tasks. MIMIC-IV-CDM is also based on MIMIC-IV but uses different task formulations and preprocessing—including filtered, denoised patient histories—and emphasizes diagnostic tasks, testing generalization to task formulation. EHRSHOT is derived from a different healthcare system with distinct event types and medical entities, serving as a benchmark to evaluate generalization on a entirely new patient ditribution. More detailed desription and case demonstration for each dataset can be found in Section \S\ref{sec:benchmarks}.

\paragraph{\textbf{\benchname}.}

To comprehensively evaluate LLM performance on EHR analysis, we introduce \benchname, a benchmark derived from MIMIC-IV~\cite{johnson2023mimic}. As summarized in Figure~\ref{fig: data overview}, \benchname spans \textbf{12 subtypes} and \textbf{42 tasks}, organized into two major groups—decision making and risk prediction—covering a broad spectrum of clinically relevant settings~\cite{DBLP:journals/corr/abs-2502-06124,DBLP:journals/npjdm/RencJSWLBS24,DBLP:conf/ml4h/GuptaGCDPB22}.

The decision-making tasks are generative, requiring the model to recommend the next appropriate intervention given a specific medical event. 
We organize these into seven subtypes—\textit{reassignment}, \textit{service}, \textit{procedure}, \textit{test \& exam}, \textit{diagnosis}, \textit{treatment}, and \textit{ICU event}—covering, for example, where a patient should be transferred (transfer), which tests to order (test \& exam), or the likely disease (diagnosis). These subtypes comprise \textbf{24 tasks} that assess the model’s ability to map a patient’s current state to clear, actionable decisions.

In contrast, risk-prediction tasks are binary classification problems in which the model forecasts whether a significant medical event will occur within a specified horizon. We group these into five subtypes—\textit{mortality}, \textit{readmission}, \textit{length of stay}, \textit{transfer}, and \textit{outcome}.
Specifically, \textit{transfer} refers to events such as a patient's admission or transfer to another department, and outcome aggregates severe events such as death or ICU transfer. 
These account for \textbf{18 tasks} that probe the model’s capacity to identify risks from longitudinal patient data. 

By spanning both generative and predictive settings, \benchname offers a comprehensive framework that mirrors real-world EHR challenges and rigorously tests LLM reasoning and adaptability in clinical contexts.

\paragraph{\textbf{MIMIC-IV-CDM}.}
The MIMIC-IV Clinical Decision Making~(MIMIC-IV-CDM)~\cite{hager_evaluation_2024} benchmark is also derived from MIMIC-IV. Compared with \benchname, its distinct patient preprocessing pipeline and task design focus enable the evaluation of models’ generalization across domain shifts in EHR analysis tasks. Specifically, it targets diagnostic accuracy for four diseases—appendicitis, cholecystitis, diverticulitis, and pancreatitis—framed as decision-making tasks. We assess two diagnostic granularities: main disease diagnosis and ICD-level full diagnosis, enabling evaluation across coarse and fine levels of clinical specificity.

\paragraph{\textbf{EHRSHOT}.}

EHRSHOT, a public dataset from Stanford Medicine, is used to evaluate models’ generalization to an entirely new healthcare system, which exhibits domain shifts not only in task formulations but also in population demographics, event type distributions, and recording practices. The benchmark comprises 14 risk-prediction tasks across three subtypes: \textit{operational outcomes}~(long length of stay, 30-day readmission, ICU transfer), \textit{anticipating lab test results}~(thrombocytopenia, hyperkalemia, hypoglycemia, hyponatremia, anemia), and \textit{assignment of new diagnoses}~(hypertension, hyperlipidemia, pancreatic cancer, celiac disease, lupus, acute MI). Following the original protocol, we assess both zero-shot performance (directly on unseen data) and few-shot adaptation given 1 to 128 examples, probing robustness and sample-efficient transfer.

\subsubsection{Metrics}

We evaluate decision-making and risk-prediction tasks with metrics aligned to their outputs, more detailed formulation can be found in Section \S\ref{section: metric}.

\paragraph{\textbf{Decision-making tasks}} are formulated as a multi-label prediction problem, where the model outputs a set of medical entities per sample. We report \textbf{F1 score}, balancing \textbf{precision} (fraction of predicted entities that are correct) and \textbf{recall} (fraction of ground-truth entities recovered). Only exact entity matches are counted as correct. To accommodate general-purpose LLMs without EHR-specific fine-tuning ({\em e.g.}, GPT-4o, DeepSeek-R1), we provide a candidate pool: 100 randomly sampled labels from the task’s full label space combined with the true labels for models to select, avoiding penalization for unfamiliar output spaces.

\paragraph{\textbf{Risk-prediction tasks}} are formulated as binary classification, which can be evaluated with \textbf{Area Under the Receiver Operating Characteristic curve~(AUROC)}. It is a robust metric that measures a model's ability to distinguish between positive and negative outcomes across all possible classification thresholds, making it ideal for medical risk prediction where datasets can often be imbalanced.

Since large language models don't have a traditional classification head, we get our probability estimates by using the tokens \texttt{yes} and \texttt{no} as our positive and negative classes, under a yes/no question prompt. At inference, we apply a technique called \texttt{logit\_biases} to isolate the model's scores for \texttt{yes} and \texttt{no}~\cite{nori2023capabilities}. We then normalize these scores using a softmax function to get a probability for the positive class, which we use to calculate AUROC.
Further metric details are provided in Section \S\ref{section: metric}.

\subsubsection{Baselines}
\label{sec: results baselines}
To provide a comprehensive performance comparison, we select a diverse type of LLMs as our baselines. More details can be found in Section \S\ref{sec:baselines}.

\vspace{-5pt}
\begin{itemize}
    \setlength\itemsep{0.5em}
    \item \textbf{Llama3.3-70B}~\cite{DBLP:journals/corr/abs-2407-21783}: Developed by Meta, this series is a family of powerful open-source models that serve as a robust foundation for a wide range of applications.
    \item \textbf{Qwen2.5-72B}~\cite{team2024qwen2}: As a versatile, general-purpose multilingual model from Alibaba Cloud, Qwen2.5 is recognized as one of the strongest open-source models available.
    \item \textbf{GPT-4o}~\cite{hurst2024gpt}: As OpenAI's flagship closed-source model, GPT-4o is a powerful multimodal model known for its advanced reasoning and real-time responsiveness.
    \item \textbf{Qwen3-235B}~\cite{yang2025qwen3}: The \textbf{reasoning model} offers a unique `thinking' and `non-thinking' hybrid reasoning engine for enhanced problem-solving.
    \item \textbf{DeepSeek R1}~\cite{DBLP:journals/corr/abs-2501-12948}: This is the largest open-source \textbf{reasoning model} we evaluate, primarily focused on pushing the limits of reinforcement learning to achieve state-of-the-art results in complex reasoning.
    \item \textbf{GPT-OSS-120B}~\cite{gptoss}: Released by OpenAI, this open-source \textbf{reasoning model} provides strong reasoning and agentic capabilities to the open community.
    \item \textbf{Medgemma-27B}~\cite{sellergren2025medgemma}: Created by Google, this is a specialized \textbf{medical model} optimized for medical text and image comprehension, with a key multimodal capability.
    \item \textbf{OpenBioLLM-70B}~\cite{dorfner2024biomedical}: An open-source \textbf{medical model} meticulously fine-tuned for the biomedical field based on the Llama3-70B, consistently outperforming larger general-purpose models on biomedical benchmarks.
    \item \textbf{Baichuan-M2-32B}~\cite{wang2025baichuan}: This is a powerful open-source \textbf{medical reasoning model} based on Qwen3-32B. It has been specifically optimized for medical scenarios and trained from scratch on an unprecedented amount of high-quality medical data, enabling it to achieve deep medical expertise.
\end{itemize}

\begin{figure}[!t]
    \centering
    \includegraphics[width=1\linewidth]{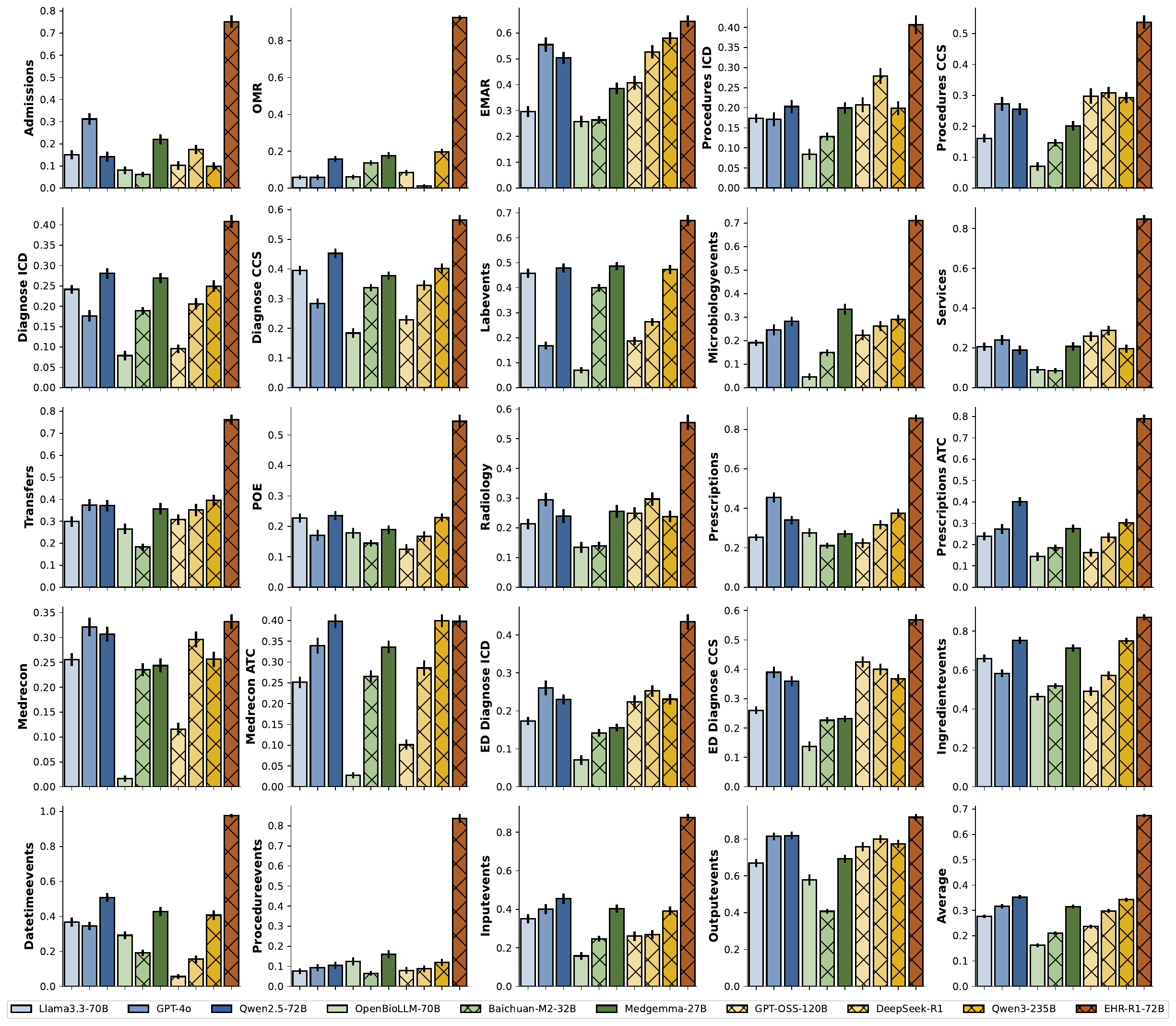}
    \caption{\textbf{Performance comparison of \modelname and nine baseline LLMs across 24 decision-making tasks on \benchname.} The performance is measured with F1 score. 
    Cross-hatched bars denote reasoning-enhanced models, highlighting the effect of explicit reasoning. 
    In each subplot, our \modelname~(rightmost bar) achieves a clear performance advantage on nearly all tasks.}
    \label{fig: decision making}
\end{figure}

\subsection{Results on \benchname}

\begin{figure}[!t]
    \centering
    \includegraphics[width=1\linewidth]{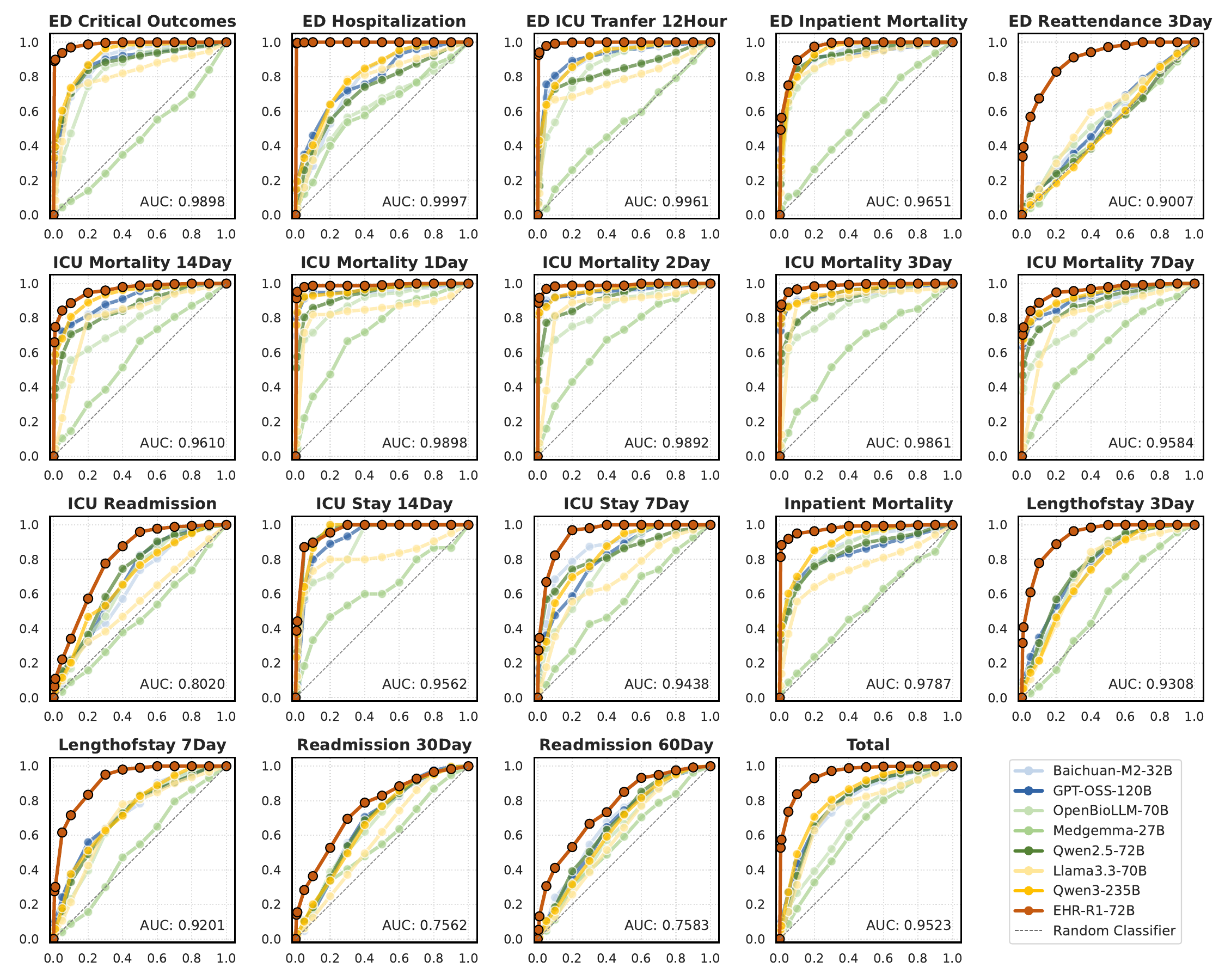}
    \caption{\textbf{Performance comparison of \modelname and seven baseline LLMs on 18 risk-prediction tasks on \benchname.} 
    Each subplot shows the ROC curves per task, with \modelname highlighted in orange; the bottom-right corner of each plot reports \modelname-72B’s AUROC. The final `Total' subplot summarizes aggregated performance across all 18 tasks.}
    \label{fig: risk prediction}
\end{figure}

As shown in Figure~\ref{fig: decision making} and \ref{fig: risk prediction}, we report the performance of \modelname on both decision-making and risk-prediction tasks, benchmarking against strong baselines to demonstrate the effectiveness across diverse EHR analyses. 

\subsubsection{Decision-making Tasks}
We show the results of evaluated LLMs with bar graph in Figure~\ref{fig: decision making} and the accurate numerical performance in the Supplementary Table~\ref{tab: deicsion making}.

\textbf{\modelname-72B demonstrates superior results.}
It achieves an average F1 of 0.6744, outperforming the next-best model, Qwen2.5-72B (0.3535), by over 30 points. 
The advantage is consistent across all decision-making tasks, validating the effectiveness and scalability of our domain-specific, reasoning-driven approach. 
\modelname-72B also surpasses strong closed-source models (GPT-4o: 0.3155) and specialized medical LLMs (Med-Gemma-27B: 0.3157), highlighting superior clinical reasoning and EHR-specific knowledge.

\textbf{Strong models perform inconsistently.} A notable observation is that closed-source commercial models do not reliably outperform open-source ones. For example, GPT-4o~(0.3155) performs the second best on average, but is often surpassed by Medgemma-27B (0.3157) and Qwen2.5-72B (0.3535) in inpatient diagnostic settings. Similarly, the general-purpose DeepSeek-R1 model (0.2974) underperformed both Medgemma-27B and Qwen2.5-72B. This variability underscores limitations in generalization for prior LLMs. In contrast, \modelname-72B delivers consistently strong F1 across diverse decision-making tasks, suggesting it can construct task-tailored reasoning pathways absent in existing models.

\textbf{General reasoning shows limited benefit.} 
Despite the common consensus that reasoning capabilities improve model performance, existing reasoning models that are not tailored to EHR tasks actually fail to demonstrate a clear advantage over non-reasoning ones. For instance, the top-performing non-reasoning model, Qwen2.5-72B (0.3535), outperforms the best-performing reasoning baseline model, Qwen3-235B (0.3418). Similarly, the non-reasoning Medgemma-27B (0.3157) achieves a better score than the reasoning-enhanced DeepSeek-R1 (0.2977). 
This indicates that effective reasoning for EHR requires tight integration of medical knowledge with case-structured analysis rather than general chain-of-thought scaling.

In summary, \modelname-72B not only sets a new state-of-the-art for decision-making tasks in EHR analysis but also addresses critical limitations of prior models by seamlessly integrating domain-specific reasoning with medical expertise. 

\subsubsection{Risk-prediction Tasks}

\textbf{\modelname-72B consistently leads}. 
Figure~\ref{fig: risk prediction} illustrates the performance of our proposed model, against several open-source baselines on 18 risk-prediction tasks, measured by ROC curves. \modelname-72B attains an average AUROC of 0.9523, significantly outperforming the second-best baseline model, Qwen3-235B (0.8245). The performance gains are especially pronounced in challenging Emergency Department~(ED) tasks, where rapid and accurate risk assessment is critical. For instance, on the ED reattendance 3day task, \modelname-72B reaches 0.9007 versus Qwen2.5-72B's 0.5540. For core clinical predictions, the model also excels—for example, inpatient mortality at 0.9787 versus Qwen3-235B’s 0.9028.

\textbf{Baseline LLMs lack task versatility.} The baselines' performance varies significantly across task types, indicating limited generalization. 
Notably, Qwen2.5-72B—strong in decision-making—trails Qwen3-235B (0.8245) on risk prediction despite similar or smaller parameter counts, and both Qwen2.5-72B and Med-Gemma-27B fluctuate across tasks. These inconsistencies suggest current general-purpose and medical LLMs struggle to deliver uniformly reliable performance across the full spectrum of EHR analysis.

\textbf{Prediction accuracy decreases with longer horizons}. 
All models exhibit decreasing AUROC as the prediction horizon extends, reflecting the complexity of forecasting long-term outcomes from longitudinal, multi-factor trajectories. For ICU Mortality, \modelname-72B drops from 0.9898 (1-day) to 0.9610 (14-day). While our model consistently outperforms baselines even on these difficult, long-range predictions—for example, scoring 0.7562 on the Readmission 30Day task against Qwen2.5-72B's 0.6735—it's clear that predicting long-term patient outcomes remains a significant challenge. 

\subsection{Generalization Evaluation on MIMIC-IV-CDM}

\begin{figure}[!t]
    \centering
    \includegraphics[width=1\linewidth]{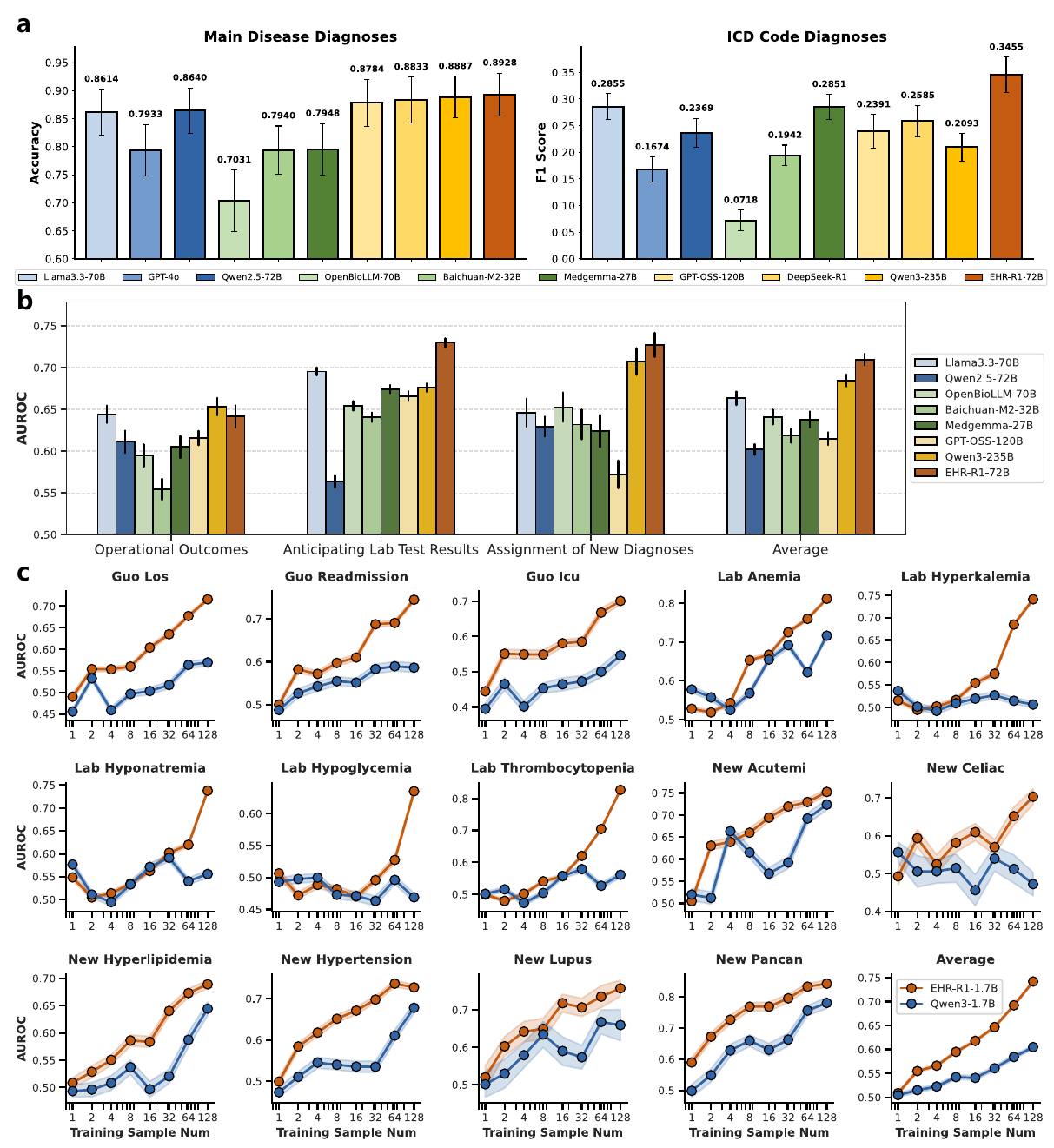}
    \caption{\textbf{Performance comparison of \modelname and baseline LLMs in generalization evaluation.} The score metric for MIMIC-IV-CDM and EHRSHOT is F1 and AUROC score, respectively. \textbf{a} Zero-shot on MIMIC-IV-CDM:
    performance on Main Disease and ICD-level diagnoses using the same samples.\textbf{b} Zero-shot on EHRSHOT for 70B-parameters LLMs: aggregated AUROC across all subtasks within each of the three category groups, plus the overall average across all tasks. \textbf{c} Few-shot on EHRSHOT for small-scale language models~(Qwen3-1.7B vs.~\modelname-1.7B): per-task performance across$k\in\{1,2,4,8,16,32,64,128\}$ shots.}
    \label{fig: ehrshot}
\end{figure}

The results of our generalization evaluation on the MIMIC-IV-CDM tasks are shown in Figure~\ref{fig: ehrshot}a. We focus on the zero-shot setting, where models are directly prompted~\cite{brown2020language} to perform EHR analysis tasks in MIMIC-IV-CDM without any task-specific training. The prompts used for evaluation are shown in Supplementary Table~\ref{fig: ood_instruction}. We evaluate LLMs' performance on two levels of diagnosis tasks: the main disease diagnoses and ICD coding level diagnoses. It can be observed that our model, \modelname-72B, achieves the highest performance on both types of tasks.

\textbf{\modelname-72B excels on main disease prediction.} 
For the task of predicting the main disease, which provides a less noisy and more idealized scenario, our model sets a new state-of-the-art with a performance of 0.8913. While other powerful models like DeepSeek-R1 (0.8841) and GPT-OSS-120B (0.8793) also perform well, our model maintains a slight lead. This demonstrates that our model's diagnostic capabilities are highly effective even in a simplified and refined clinical context.

\textbf{\modelname-72B leads in multi-level diagnoses.} 
Most models struggle to perform well on both main disease and ICD-level coding simultaneously. For example, DeepSeek-R1 (0.8841 Main Disease) and GPT-OSS-120B (0.8793) lag on ICD Code prediction (0.2597 and 0.2422), whereas Med-Gemma-27B attains higher ICD accuracy (0.2860) but lower Main Disease (0.7939).
In contrast, \modelname-72B, is the only one that achieves the highest performance on both tasks, with scores of 0.8913 for main disease and 0.3501 for ICD code. These results underscore its ability to generalize zero-shot to a new system while retaining granular coding precision.

\subsection{Generalization Evaluation on EHRSHOT}

We follow the EHRSHOT protocol to assess generalization in both zero-shot and few-shot scenarios: (1) zero-shot, where we directly evaluate on an unseen dataset with prompts to measure out-of-the-box generalization similar as n MIMIC-IV-CDM, and (2) few-shot, where we \textbf{train} on a small number of labeled examples to gauge how quickly LLMs adapt with minimal supervision. The prompts used for evaluation are shown in Supplementary Table~\ref{fig: ood_instruction}.

\subsubsection{Zero-shot Setting}

In the zero-shot setting~(Figure~\ref{fig: ehrshot}b), 
we compare \modelname-72B to similarly sized (~70B) baselines across three categories—operational outcomes, lab test forecasting, and new-diagnosis assignment—and the overall average. \modelname-72B attains the highest AUROC in every category and the overall average, outperforming strong general models ({\em e.g.}, Qwen2.5-72B) and specialized medical models ({\em e.g.}, Med-Gemma-27B). Given that EHRSHOT differs from \benchname in both data format and task types, these results underscore robust cross-dataset generalization, which is a critical feature for practical clinical applications.

\subsubsection{Few-shot Setting}
Considering that model adaptation in clinical settings is often constrained by limited computational resources. Therefore, in the few-shot setting, we use a small-parameter 1.7B LLMs with a minimal number of training samples~(at most 128) to evaluate the performance of the LLMs in rapid adaptation scenarios.

\textbf{\modelname shows better learning efficiency.} Figure~\ref{fig: ehrshot}c and the accompanying table present a detailed performance comparison between our small model, \modelname-1.7B, and its base model, Qwen3-1.7B, on a variety of EHRSHOT tasks under a few-shot fine-tuning setting. Following EHRSHOT, the models are fine-tuned on a limited number of examples ($k$) for each task, ranging from 1 to 128. The results show the profound superiority of our model, which achieves an average AUROC of 0.7465 at $k=128$, far exceeding Qwen3-1.7B's average of 0.5998 at the same shot setting.

\textbf{Performance gap grows with more data.} 
Our model's performance advantages are primarily manifested in two types of tasks: operational outcomes and lab test forecasting. When $k<16$, the gap is modest—likely because extremely limited data under-utilizes large models. For $k \geq 16$, the gap widens markedly, suggesting that EHR-focused training equips \modelname-1.7B to leverage small-to-moderate datasets more efficiently.

\textbf{General models narrow the gap on diagnosis tasks.} 
For new-diagnosis prediction, the gap peaks around $k=16$ ({\em e.g.}, new hypertension: 0.6593 vs.~0.5108 AUROC for \modelname-1.7B vs. Qwen3-1.7B) but narrows as k increases. This likely reflects that disease diagnosis is a core medical task extensively represented in general-model pretraining; with sufficient fine-tuning data, general models can catch up.

In summary, our model not only achieves higher performance but also improves faster as data scales, indicating superior learning efficiency and a stronger EHR-specific knowledge base.

\begin{figure}[!t]
    \centering
    \includegraphics[width=1\linewidth]{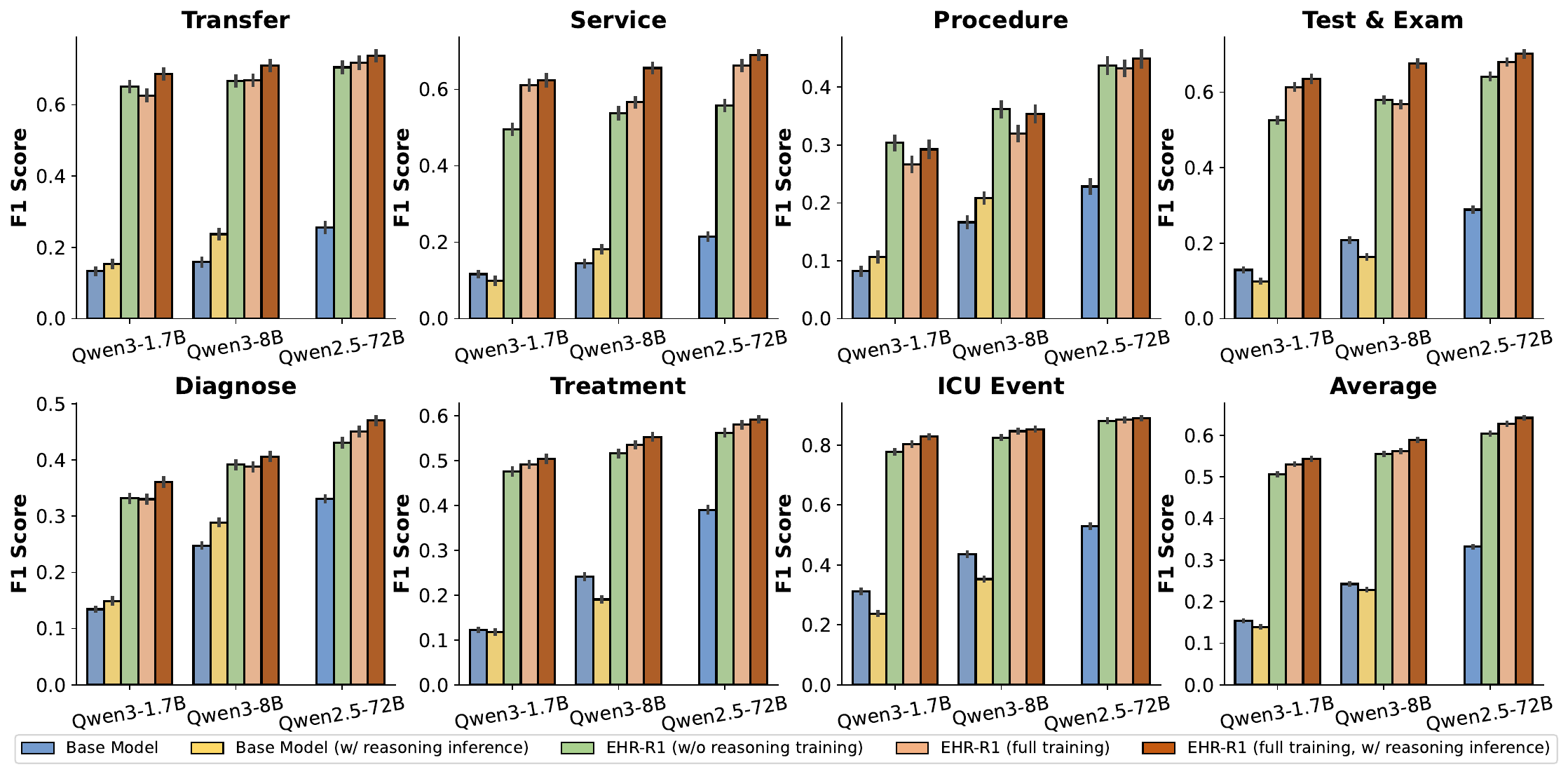}
    \caption{\textbf{Ablation experiments results on decision making tasks of \benchname.} The figure show 7 categories of sub-type decision making tasks and the average performance of 5 variant of our methods, including Base Model, Base Model (with reasoning inference), EHR-R1 (w/o reasoning training), EHR-R1 (full training), and our final model EHR-R1 (full training, w/ reasoning inference), which showcase the incremental performance gains from each stage of our training pipeline.}
    \label{fig: ablation}
\end{figure}

\subsection{Ablation Study}
We conduct ablation experiments to validate the rationale and effectiveness of our proposed framework from \textbf{three key perspectives}: the impact of incorporating reasoning data into model training, the effectiveness of reasoning during inference compared to direct answer output, and the generality of our approach across model scales.

\paragraph{Experiments design.} 
We conducted an ablation analysis on five configurations, 
progressively adding our training and inference components~(Figure~\ref{fig: ablation}): (i) \textbf{BaseModel}: the original base model with direct answers~(no reasoning at inference, no training enhancements). (ii) \textbf{BaseModel (w/ reasoning inference)}: the base model evaluated with test-time reasoning to guide predictions. (iii) \textbf{\modelname (w/o reasoning training)}: the model has undergone our continual pre-training to specialize in the EHR domain, but without reasoning data; inference uses direct answers. (iv) \textbf{\modelname (full training)}: our three-stage model—continual pre-training, reasoning-data training, and task-specific fine-tuning—with direct-answer inference. 
(v) \textbf{\modelname (full training, w/ reasoning inference))}: the full model augmented with test-time reasoning.
Note that, Qwen2.5-72B is not applied with \textbf{BaseModel (w/ reasoning inference)} because it is not a reasoning-enabled model. The accurate experimental results can be found in the Supplementary Table~\ref{tab: ablation}.

\paragraph{Effectiveness of reasoning data.} 
Incorporating reasoning data into training consistently improves performance: for \modelname-1.7B, direct-answer F1 rises from 0.5060 (without reasoning data) to 0.5300 (with reasoning data), and for \modelname-72B from 0.6039 to 0.6281, indicating that synthesized reasoning injects useful EHR-specific knowledge. The gains are amplified when reasoning is also used at inference, with the full \modelname-1.7B reaching 0.5438 F1 compared to 0.5060 without reasoning data, demonstrating that training-time reasoning and test-time reasoning are complementary—training equips the model with reasoning primitives, and inference-time reasoning leverages them to further refine predictions—yielding a more knowledgeable and robust foundation across inference strategies. 

\paragraph{Effectiveness of reasoning inference.}
Reasoning at inference only helps when the model has been explicitly equipped with EHR-specific reasoning. Applying general reasoning prompts to base models offers little-to-negative benefit (Qwen3-1.7B drops from 0.1624 to 0.1456 F1; Qwen3-8B also drops from 0.2425 to 0.2286), underscoring that general reasoning alone is insufficient for clinical tasks. In contrast, our fully trained EHR-R1 models gain substantially from reasoning inference: EHR-R1-1.7B improves from 0.5060 to 0.5438 ($\pm0.0035$), EHR-R1-8B from 0.5549 to 0.5894, and EHR-R1-72B from 0.6039 to 0.6418. These results validate the effectiveness of our `thinking-graph' synthesis pipeline in instilling domain-specific reasoning pathways that inference-time reasoning can reliably exploit.

\paragraph{Effectiveness across model scales.}
Our approach generalizes robustly with scale: applying the same pipeline from 1.7B to 72B parameters yields monotonic gains, with average F1 rising from 0.5438 (Qwen3-1.7B–based) to 0.5894 (Qwen3-8B–based) and 0.6481 (Qwen2.5-72B–based). This consistent improvement indicates that the benefits of reasoning-data training and reasoning inference are not scale-specific but instead compound as model capacity increases, confirming that our framework scales effectively across parameter regimes.
\section{Discussion}

Despite the remarkable progress of LLMs across diverse medical tasks, their application to clinical EHR analysis remains a significant challenge~\cite{rouhizadeh2025large, qu2024enhancing, soroush2024large, ren2025comprehensive}. Existing approaches fall short in two critical dimensions. First, task coverage remains narrow, as most efforts are confined to specific objectives or disease cohorts rather than enabling holistic analytical support for evolving clinical workflows. Second, reasoning ability is underdeveloped, with current LLMs struggling to generate reliable EHR-focused reasoning processes that progressively highlight key records, integrate fragmented evidence, and construct longitudinal, synthesized understanding of patient health.



\textbf{Overview of Our Approach}

We introduce a thinking-graph–driven, auto-generation pipeline for EHR reasoning that systematically converts raw, longitudinal records into structured, query-ready insights for generative LLM training. The pipeline: (1) extracts salient medical entities from heterogeneous EHR sources to create clinical focused insights; (2) links these entities into a thinking graph that encodes temporal relations and causal hypotheses, transforming disjoint timesteps into a coherent longitudinal narrative; and (3) synthesizes explicit, stepwise reasoning over this graph to support adaptive solutions across diverse clinical queries. This pipeline enables to curate \textbf{\traindataname}, 
a super-instruction corpus with 300K reasoning cases and 3.5M non-reasoning cases spanning 42 EHR tasks. We then train our reasoning-enhanced model, \textbf{\modelname}, via three-stage training, including large-scale domain adaptation, reasoning enhancement followed by reinforcement learning with Group Relative Policy Optimization (GRPO) to further strengthen analytical fidelity, temporal reasoning, and clinical robustness.

\textbf{Main Contribution}

\textbf{A super-instruction dataset for EHR analysis with reliable reasoning.} 
We release EHR-Ins, a large-scale super-instruction corpus that pairs 300K high-quality reasoning cases with 3.5M non-reasoning cases across 42 EHR tasks. Reasoning traces are auto-generated via our thinking-graph pipeline, which extracts entities, links temporal/causal relations, and synthesizes clear stepwise rationales—transforming structured EHR data into LLM-ready, actionable insights. This design enables models to both answer diverse clinical queries and explain their decisions.

\textbf{First comprehensive benchmark for EHR analysis.} 
This paper introduces EHR-Bench, a new benchmark based on the MIMIC-IV dataset, which provides a comprehensive evaluation of LLMs on 42 diverse EHR tasks. This benchmark is designed to holistically assess a model's ability to handle the full range of diverse clinical queries that exist in real-world EHRs, addressing coverage gaps in existing evaluations.

\textbf{A robust and generalizable foundation model.} We introduce \modelname, a state-of-the-art model that demonstrates superior performance on EHR analysis. By learning from the explicit reasoning paths in our data, our model not only sets new performance benchmarks on the proposed EHR-Bench but also shows exceptional zero-shot generalization capabilities on the out-of-distribution EHRSHOT dataset. This highlights \modelname's robust ability to translate EHR content into clinical insight and understand temporal data to form a longitudinal narrative, which is critical for real-world clinical applications.

\textbf{Key Findings}

\textbf{Effective handling of diverse tasks.} 
Trained on our EHR-Ins dataset, \modelname performs a broad range of EHR tasks with high accuracy. Across 42 tasks in EHR-Bench—including clinical coding and other less-common or challenging tasks—it delivers strong, adaptive reasoning over diverse clinical queries. In addition, its zero-shot performance on unseen EHR data and tasks from the EHRSHOT dataset confirms that the model captures transferable structure rather than memorizing training data.

\textbf{Effectiveness of reasoning for EHR tasks.} 
The gains derive from an auto-synthesis pipeline that leverages relationships among medical entities to teach explicit reasoning over structured EHR data. This enables the model to denoise heterogeneous inputs, analyze temporal relations, and construct longitudinal clinical narratives. Applying reasoning chains at inference yields substantial improvements, supporting the value of explicit, knowledge-informed reasoning.

\textbf{Generalizability of our framework.} 
The effectiveness of our method is not limited to a specific model scale. Improvements from our training framework and reasoning data are consistent for models from 1.7B to 72B parameters. This finding demonstrates the generalizability and scalability of our approach, confirming that it provides a robust and widely applicable solution for the EHR domain.

\textbf{Limitations and Future Work}

\textbf{Scope of reasoning data.} 
Our reasoning-data synthesis was applied only to decision-making tasks, for which target labels include the medical entities necessary for building the thinking graph. This constrains coverage of diverse clinical queries. Although stage-three reinforcement learning transferred some of these capabilities to risk-prediction tasks, future work should explore procedures to construct explicit reasoning data for these binary classification.

\textbf{Data attrition during graph construction.} 
We excluded many samples for which the thinking graph could not be built—either due to insufficiently related entity pairs or unsuccessful retrieval of medical relations from UMLS. 
This limits our ability to convert temporal data into a longitudinal narrative for every case. Refining graph-building heuristics, incorporating additional or domain-specific knowledge bases, and improving entity linking could reduce data loss and broaden case coverage.

\textbf{Specialization versus breadth}. 
The approach yields a specialized EHR model and may attenuate general-purpose capabilities. Emphasizing structured temporal narratives and prescribed reasoning steps risks overfitting to clinical settings. Future work should explore strategies to balance specialization and breadth—for example, multi-domain continued pretraining, alternating-task curricula, modular adapters, or mixture-of-experts routing—to preserve general capabilities while deepening clinical expertise.

\section{Methods}
In this section, we present the details of our method, starting with the problem formulation, followed by the reasoning data curation pipeline, model training, and finally, evaluation.

\subsection{Problem Formulation}


We formulate the EHR analysis problems in an instruction-tuning-based~\cite{ouyang2022training} generative framework.  

An EHR can be represented as a chronologically ordered sequence of longitudinal records (also referred to as events), including laboratory results, medication administrations, transfers, \emph{etc.}:  
\begin{equation}
    \mathcal{R} = \{\mathbf{r}_1, \mathbf{r}_2, \dots, \mathbf{r}_{K}\}.
\end{equation}
The $k$-th record is defined as $\mathbf{r}_k = (c_k, E_k, t_k)$, where $c_k \in \mathcal{C}$ denotes the clinical event category and $\mathcal{C}$ represents the set of all possible event types, such as \textit{medication}, \textit{diagnosis}, \textit{laboratory test}, \emph{etc.}. The term $E_k$ represents the set of medical entities contained in the event (e.g., test items and results in a laboratory event, or diagnoses in a diagnostic event). Finally, $t_k$ is the timestamp of the event.

At any given prediction time step $t$, a training sample for EHR analysis is expressed as
\[
    S = (\mathcal{I}, \mathcal{R}_{\leq t}, \mathcal{A}),
\]
where $\mathcal{I}$ is a free-text instruction specifying the analysis task, $\mathcal{R}_{\leq t} = \{\mathbf{r}_k \mid t_k \leq t\}$ is the observable history of the EHR up to time $t$, and $\mathcal{A}$ is the ground-truth free-text answer.  

Our objective is to train an LLM, $\Phi_\text{LLM}$, to perform the following conditional text generation task:
\begin{equation}
    \Phi_{\text{LLM}}(\mathcal{I}, \mathcal{R}_{\leq t}) \;\rightarrow\; \mathcal{A},
\end{equation}
that is, to output an appropriate answer $\mathcal{A}$ given the task instructions $\mathcal{I}$ and the observed EHR history $\mathcal{R}_{\leq t}$.

\begin{figure}[!t]
    \centering
    \includegraphics[width=.9\linewidth]{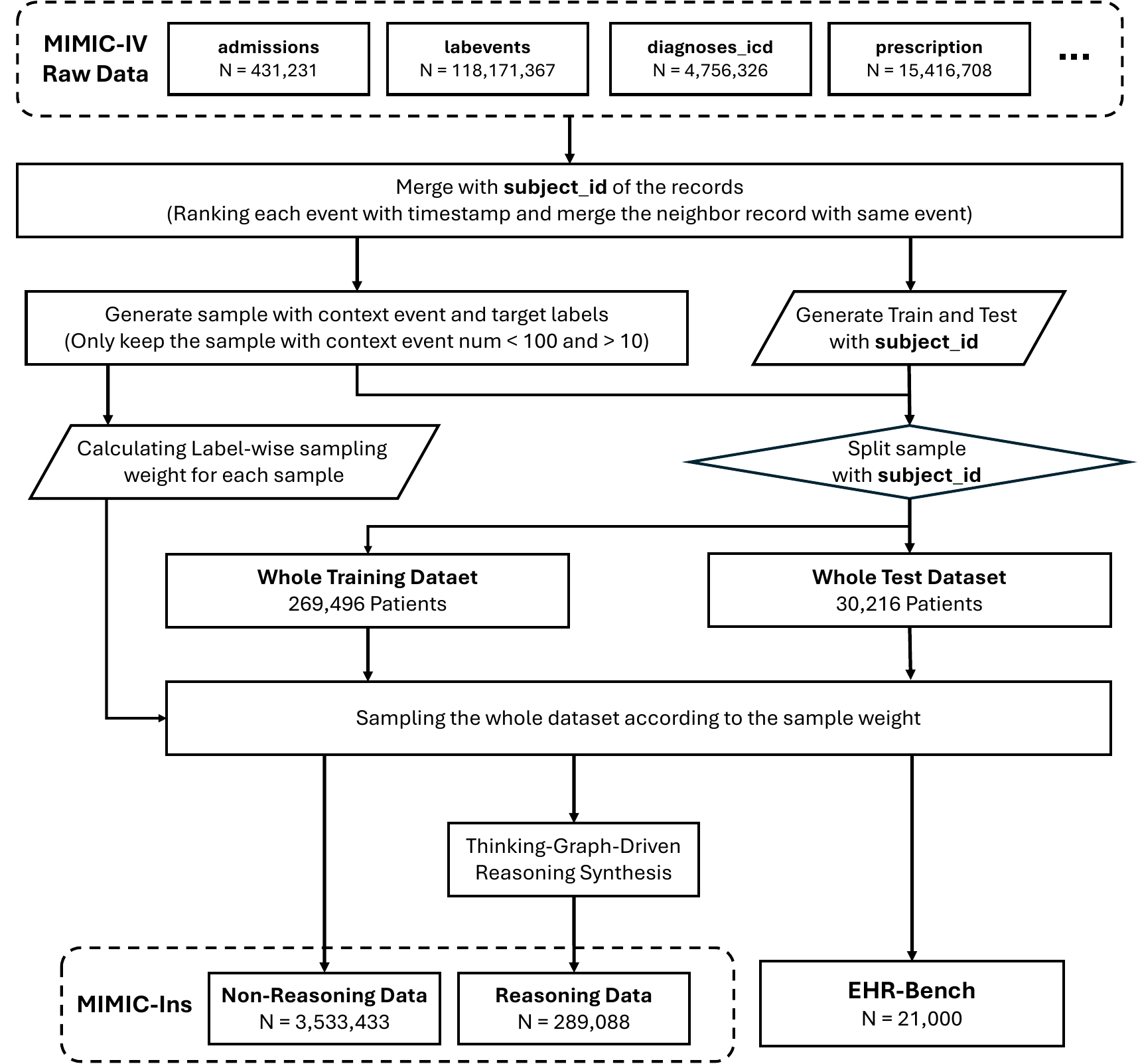}
    \vspace{8pt}
    \caption{\textbf{Overview of Data curation pipeline.} The pipeline for our data curation process begins with the original MIMIC-IV dataset. From there, we establish EHR-Ins with reasoning-enhanced EHR analysis instructions through thinking-graph–driven reasoning synthesis, along with a new comprehensive EHR-Bench. }
    \label{fig: data curation}
\end{figure}

\subsection{Data Curation}

In this section, we introduce the curation pipeline of \textbf{\traindataname} and \textbf{ \benchname}. 
We begin by processing the raw MIMIC-IV EHR cases, filtering and reorganizing them into an instruction-tuning generative format. The data is then carefully split into training and testing sets, where the test set forms \benchname, and the training set is further enhanced with reasoning, resulting in the final \traindataname. The overview of the preprocessing pipeline is shown in Figure~\ref{fig: data curation}.

\subsubsection{MIMIC-IV Processing}

Here, we describe the construction process of both \traindataname and \benchname. First, we introduce the processing details of MIMIC-IV. Then, we describe the free-text formatting of the EHR data, which transforms the long-horizon EHRs into detailed generative training samples.  Finally, we describe the approach to achieve a balanced distribution of the data with label-wise weighted sampling methods and how to split the whole dataset into a training and a test set.

\paragraph{Information Enrichment.} To enable chronological construction from EHR data, we reformat and sort the original MIMIC-IV dataset at patient level. We first extract all events for each patient and reordered them by timestamp with second precision. Recognizing that \textit{diagnoses\underline{ }icd}, \textit{procedure\underline{ }icd}, and \textit{diagnosis} event types in MIMIC-IV lack second-level timestamps, we manually added them to enrich the information. In particular, for \textit{diagnoses\underline{ }icd} and \textit{diagnosis}, we associate them with their respective admission events via \textit{hadm\underline{ }id}. We then set \textit{diagnoses\underline{ }icd} to 1 minute before the \textit{discharge} event and \textit{diagnosis} to 1 minute before the emergency department discharge event. For \textit{procedure\underline{ }icd} events, which only provide day-level timestamps, we default their time to 23:59:59 on that day. 

To further enrich the EHR information, we map the International Classification of Diseases~(ICD) codes in the \textit{diagnoses\underline{ }icd} and \textit{procedure\underline{ }icd} tables to Clinical Classifications Software (CCS) categories using the `ICD-to-CCS' script\footnote{ICD9 to CCS: \url{https://hcup-us.ahrq.gov/toolssoftware/ccs/ccs.jsp}\\ICD10 to CCSR: \url{https://hcup-us.ahrq.gov/toolssoftware/ccsr/dxccsr.jsp}}. Likewise, we translated National Drug Codes~(NDCs) in the \textit{prescriptions} and \textit{pharmacy} tables into Anatomical Therapeutic Chemical (ATC) classification codes adopting the off-the-shelf mapping script\footnote{NDC to ATC: \url{https://github.com/sunlabuiuc/PyHealth}}. These enrichment consolidate the originally vast and heterogeneous sets of disease, procedure, and medication codes into clinically coherent categories. We also identify that information within discharge events, such as social history and chief complaint, as being observable upon admission. We move these details to the admission event to provide more comprehensive input for LLMs. Additionally, to mitigate potential data leakage in EHR data, we mask \textit{pharmacy} event information within \textit{prescription} events.

After the above processing and enrichment, all MIMIC-IV data are organized into a chronological sequence of records for each patient.

\paragraph{Training Sample Formatting}
Using the structured EHRs, we construct an instruction dataset on EHR analysis by defining detailed instruction–answer pairs for each training sample, enabling generative training, regarding the two EHR analysis task categories:  \textbf{decision-making} and \textbf{risk-prediction}.

Decision-making tasks involve predicting the next-step event based on the history of events. Therefore, naturally, for a given EHR sequence $\mathcal{R} = \{ \mathbf{r}_1, \mathbf{r}_2, \dots, \mathbf{r}_{K} \}$, we can generate training samples by sampling an arbitrary timestep $t$ and predicting the next event.
Assuming the next-step record as $\mathbf{r}_{t+1} = (c_{t+1}, E_{t+1}, t_{t+1})$, we formulate the instruction $\mathcal{I}$ for decision-making task according to the type of next-step event: $c_{k+1}$~(detailed prompts can be found in Supplementary Table \ref{fig: instruction}), 
and the target answer set is composed of the entities contained in $E_{k+1}$:
\begin{equation}
    \mathcal{A} =\{y_1,y_2,\dots,y_N\}, \text{ where } \ y_i\in E_{k+1}, \ i\in\{1,\cdots,N\},
\end{equation}
For example, in a case where the next-step event is diagnoses, $E_{k+1}$ contains all the diagnosis results of the patients and $y_i$ represents the name of a disease. In the implementation, $\mathcal{A}$ is formatted as free-text by concatenating the entity list with \texttt{str.join} function in Python.

In addition, at the time step $t$, we can also construct risk-prediction samples. We organize the instruction of the risk-prediction taks by filling $\{c_\text{critic}, T\}$ into a predefined template~(details can be found in Supplementary Table \ref{fig: instruction}), where the objective of the model is to determine whether a critical outcome event of type $c_\text{critic}$ will occur within a specified time frame $T$. Therefore, the binary ground truth answer of the task can be formulated as:
\begin{equation}
   \mathcal{A} =
        \begin{cases}
        \text{yes}, & \text{if there exists } i \in \{t+1,\dots,K\} \text{ such that } c_i=c_\text{critic} \text{ and } t_i < T, \\[0.8em]
        \text{no}, & \text{otherwise}.
        \end{cases}
\end{equation}

Thus far, we have organized training samples in the form $S = (\mathcal{I}, R_{\leq t}, \mathcal{A})$, encompassing EHR analysis tasks that fall into two main categories: decision-making and risk-prediction. \textbf{Notably}, since our instructions are generated through prompt filling, all training samples associated with a specific task (e.g., diagnostic decision-making) share the same instruction. Hence, in the following, $\mathcal{I}$ can also be used to denote a specific task.


Lastly, to convert the tabular observed EHR data $R_{\leq t}$ into a format that LLMs can process, we serialize the structured data\cite{liutapex,li2024table} into Markdown format\cite{hegselmann2025large}, which is widely present in LLM pre-training corpora, facilitating better model comprehension. Specifically, each event is represented as plain text with a title (event name and standardized start time) and content. The content uses bullet-point syntax for events with single records and table syntax for others with multiple records, as shown in Case~\ref{case: single item} and Case~\ref{case: multiple item}, respectively. The free-text case of EHR is shown in Case~\ref{case: mimic-bench}.

\paragraph{Label-wise Weighted Sampling.}
Because the constructed dataset over all valid EHR time steps contains a large number of samples and the target candidate sets in MIMIC-IV are highly imbalanced, we apply sampling to form the final training and test sets. We apply weighted sampling to rebalance the target label distribution in the training set. In practice, for decision-making tasks, given a certain task prompted by instruction $\mathcal{I}$, we first count the frequency of all possible label entities $y$ appeared in the target answer $\mathcal{A}$ and use the reciprocal of the frequency as the sampling weight. Due to that, each training sample's answer label is a list of entities, and we assign a weight based on the average of its constituent labels. Formally, for a sample $S$ with ground-truth answer as $A=\{ y_1, y_2, \dots, y_N \}$, where $N$ is the size of the label set, its weight $w_S$ is given by:
\begin{equation}
    w_S = \frac{1}{N}\sum^N_{i=1}\frac{1}{\text{Count}_{\mathcal{I}}(y_i)}, 
\end{equation}
where $\text{Count}_{\mathcal{I}}(y_i)$ represents the frequency of label entity $y_i$ in a same type of task. This strategy can also be applied to risk-prediction tasks, where the entities in answering contain only two binary items (`yes' or `no', thus $N$=1). The weighting formula, therefore, simplifies to the reciprocal of the frequency of the sample's single label, rebalancing the positive and negative classes.

The label-wise weighted sampling strategy is adopted to generate the train and test sets simultaneously. This approach ensures that the \traindataname maintains a balanced distribution of target labels while addressing the issue of class imbalance. Besides, the test set can also cover diverse types of target labels, which makes the \benchname evaluating the performance of the models more comprehensive.

\paragraph{Data Split.} Prior studies commonly treat each hospital admission (which may come from the same patient) as an independent sample for data splitting. However, in our setting, to prevent potential information leakage between the training and test sets, we adopted a rigorous patient-level data split. This approach ensures that no patient, identified by their unique \texttt{subject\_id}, appears in both the training and test sets.

Furthermore, due to context length constraints and computational resources, we limit the observable historical time window to 24 hours. To ensure an appropriate length for historical event trajectories, we filter out samples with fewer than 10 or more than 100 historical events. This process prevents issues of insufficient information or excessively long inputs.

From the entire training dataset, we sample cases using a label-wise weighting scheme to ensure balanced representation across all tasks. This process yielded 3.5M samples designated as non-reasoning data and an additional 300K samples specifically for synthesizing reasoning data. These two sets collectively form our proposed super-instruction dataset, \traindataname. Similarly, a test set of 21K samples was created from the whole test set using the same label-wise sampling approach to form our benchmark, \benchname.



\begin{figure}[!t]
    \centering
    \includegraphics[width=1\linewidth]{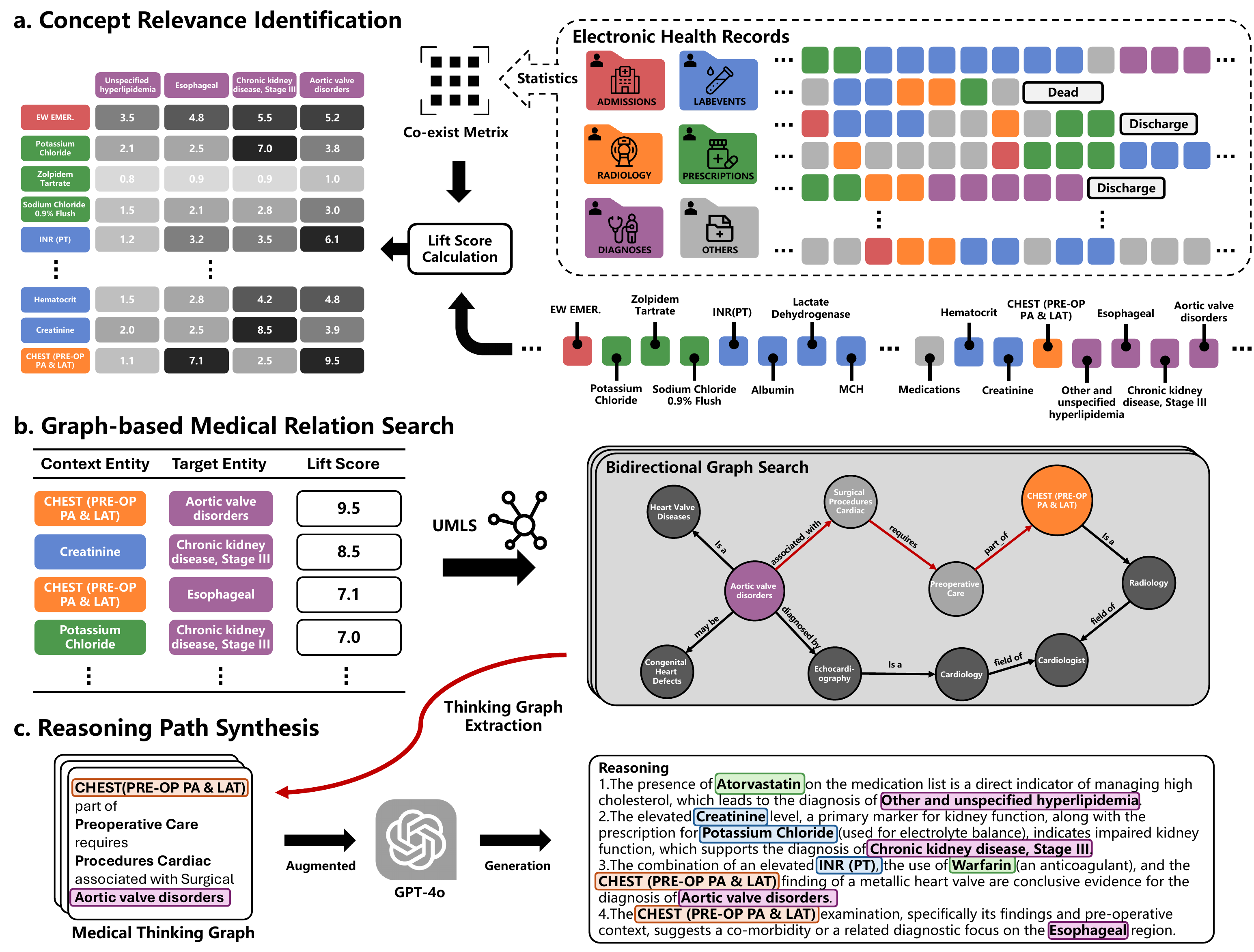}
    \vspace{1pt}
    \caption{\textbf{Overview of Thinking-Graph-Driven Reasoning Synthesis.} The whole pipeline comprises three parts: \textbf{a Entity Relevance Identification} analyzes electronic health records to calculate a co-exist matrix and Lift Score for medical entities. \textbf{b Graph-based Medical Relation Search} filters the entity pair with a higher lift score and conducts a bidirectional graph search on a large-scale medical knowledge graph, UMLS; The red lines indicate the search paths \textbf{c Reasoning Path Synthesis} arguments GPT-4o with retrieved medical relations to generate a coherent reasoning path. The entity with the color box in the reasoning chain indicates the context entity or the target labels.}
    \label{fig: reasoning synthesis}
\end{figure}

\subsubsection{Thinking-Graph-Driven Reasoning Synthesis}
\label{section: reasoning synthesis}


In this section, to enable the model to provide convincing reasoning in clinical scenarios, we further enhance the answers in \traindataname with detailed rationales by incorporating the concept distribution in the EHR and external medical knowledge sources.

We mainly consider the decision-making tasks. We first explain how to identify medical entities that are potentially related to the task labels. We then filter these entities and adopt bidirectional graph searching on external medical knowledge bases to gather the thinking graph, which links the context entities and target entities with medical relations. Finally, we describe how to synthesize the reasoning subset for decision-making tasks in \traindataname. The overview of the proposed reasoning synthesis pipeline is shown in Figure~\ref{fig: reasoning synthesis}.

\paragraph{Entity Relevance Identification.} 

Compared with regular medical data, a distinctive feature of EHRs is their tabular format, where most records are clearly indexed by a list of well-defined medical entities. Therefore, by analyzing the relationships among these entities across the training set, we can uncover potential reasoning pathways that trace key observed entities from noisy raw EHRs to task labels.

In particular, we first process some unstructured exception free-text records, such as discharge summaries and examination reports. We adopt 
\texttt{QuickUMLS}\footnote{\url{https://github.com/Georgetown-IR-Lab/QuickUMLS}} to extract the meaningful medical entities from the raw texts. Then, all the medical entities in the observable events can be expressed as $\mathcal{E}_{\le t} = \bigcup_{t_i<t}E_i$, $E_i = \{e^i_1, e^i_2, \dots, e^i_k\}$ is the entity set of the $i$-th events. 

Then we count the co-occurrence frequencies of the context entity $e_i\in\mathcal{E}_{\le t}$ with the target label entities $y_j\in \mathcal{A}$ for a specific decision-making task denoted by its instruction $\mathcal{I}$ and calculate the \textbf{Lift}~\cite{agrawal1993mining} between them. The Lift metric can discover strong associations within the data by measuring how much more frequently two entities appear together than expected by chance, allowing us to pinpoint the most relevant entities for the reasoning pathways. In our method, the Lift metric is calculated as:
\begin{equation}
    \text{Lift}(e_i, y_j, \mathcal{I}) = \frac{P_\mathcal{I}(e_i, y_j)}{P_\mathcal{I}(e_i)P_\mathcal{I}(y_j)} = \frac{\text{Count}_\mathcal{I}(e_i,y_j)\text{Count}_\mathcal{I}}{\text{Count}_\mathcal{I}(e_i)\text{Count}_\mathcal{I}(y_j)}, \ e_i\in\mathcal{E}_{\le t},\ y_j\in \mathcal{A}
\end{equation}
where $\text{Count}_\mathcal{I}(e_i,y_j)$ represents the co-occurrence frequency of $e_i$ and $y_j$ in the training set for task $\mathcal{I}$ and $\text{Count}_\mathcal{I}$ represents the total number of samples in task $\mathcal{I}$. Since rare occurrences of $e_i$ and $y_j$ (for example, occurring only once) can disproportionately inflate the \texttt{Lift} score, we only retain co-occurrence pairs where both the frequency of $e_i$ and $y_j$ are greater than 5. Finally, we retain only the co-occurrence pairs where the \texttt{Lift} value is greater than 5 for each sample:
\begin{equation}
    \mathcal{K}_{\text{Lift}} = \{ (e_i, y_j) \mid \text{Count}_\mathcal{I}(e_i)>5, \text{Count}_\mathcal{I}(y_j)>5, \text{Lift}(e_i, y_j, \mathcal{I})>5 \}, e_i\in\mathcal{E}_{\le t},y_j\in \mathcal{A}
\end{equation}

This filtering process helps to reduce noise while ensuring that the remaining entity-label pairs have a strong potential association, thus providing candidate references for evidence in the subsequent reasoning process.

\paragraph{Graph-based Entity Relation Search.}

Although Lift can uncover potentially related medical entities from the historical EHR events, the medical relationships between these entities~(including both the context entities in input and the label entities) and the labels remain uncovered and thus fail to capture the underlying clinical associations. To address this issue, we leverage the Unified Medical Language System (UMLS)~\cite{bodenreider2004unified} knowledge graph, which provides a structured representation of medical concepts and their relationships. UMLS integrates a wide range of medical terminologies and ontologies, including SNOMED CT, ICD, and MeSH, among others, making it an ideal resource to validate and strengthen the medical relationships between entities and labels in EHR data. Specifically, we map all medical entities identified in the EHR to their corresponding UMLS concepts. To increase the accuracy and completeness of this mapping, we use fuzzy matching provided in UMLS~\cite{bodenreider2004unified}, allowing us to match medical terms in the EHR with UMLS concepts even when there are slight variations in spelling, abbreviations, or synonyms. This approach ensures a broader and more robust set of entity-to-concept mappings, improving the coverage of medical entities in the EHR data. For entities that cannot be mapped to UMLS concepts, we discard the related co-occurrence pairs, as these entities would not contribute valid medical relationships and could introduce noise into the analysis. 

After mapping entities to UMLS, we search for medical relations between entity pairs, progressively linking context input entities towards the final label entities, compensating the missed concept nodes, forming a instance specific \textbf{\textit{thinking-graph}}. However, many medical entities in UMLS do not have direct associations, which makes it challenging to acquire their clinical relationships. To overcome this, we employ a bidirectional graph search algorithm to find connections between entity pairs by traversing across multiple nodes and links. This process allows us to uncover implicit medical relationships between entities that are not directly linked in the UMLS knowledge graph. 

The final discovered relationships between the entities are then extracted to form the  \textbf{\textit{thinking-graph}} to enhance the subsequent synthesis of reasoning paths for this training case:
\begin{equation}
    \mathcal{K}_{\text{Think-Graph}} = \{\text{Path}(e_i, y_j) \mid \text{Path}(e_i, y_j)\in \texttt{UMLS}, (e_i, y_j) \in \mathcal{K}_{\text{Lift}}\}
\end{equation}
where the function $\text{Path}(e_i, y_j)$ is the connection between $e_i$ and $y_j$ extracted from UMLS.

\paragraph{Reasoning Path Synthesis}
With the relationships between label entities and historical event entities for each sample, we leverage a powerful large language model, GPT-4o, to generate a detailed reasoning chain based on the historical events, entity relationships, and label entities. The prompt used for the reasoning synthesis are shown in Prompt~\ref{prompt: reasoning}.






Specifically, the reasoning chain consists of three stages: extraction, reasoning, and final result. In the extraction stage, the model is required to extract relevant information from the historical events based on the task’s objective. This extraction process must specify the events and their corresponding timestamps to clearly identify the source of the information, ensuring the accuracy and traceability of the extracted data. The reasoning stage requires the model to integrate the extracted information and, using the entity relationship data, link the extracted entities to the corresponding label entities. The model then constructs a coherent reasoning path that logically connects the historical events to the predicted label, providing an explanation for the clinical decision. In the final result stage, the model outputs the predicted result based on the reasoning path constructed in the previous stage. This output represents the model’s interpretation of the task based on the historical data and reasoning process.

It is important to note that each sample may contain multiple label entities. For those label entities for which the model cannot construct a valid reasoning chain, the synthesized data may introduce hallucinations. To address this issue, we instruct GPT-4o to generate reasoning chains only for the label entities that are logically inferable from the available data. For those label entities that cannot be reasoned through, we discard the associated reasoning chains and results. Finally, we retain only those samples where the number of inferable label entities constitutes at least 70\% of the original label entities. This threshold ensures that the data distribution remains close to the original raw data and minimizes the potential bias introduced by the discarded entities. Figure~\ref{fig: case} shows an example of the reasoning synthesis process.

\subsection{Model Training}
\label{sec:model_training}

In this section, we detail the training procedure. The proposed model follows a three-stage training process: large-scale domain adaptation to learn the data distribution and task knowledge in the EHR, reasoning enhancement to extract key information from observable historical events and infer prediction results, and reinforcement learning to enhance the model’s reasoning capabilities on EHR data. 

We first serialize the training sample $S=(\mathcal{I},\mathcal{R}_{\leq t}, \mathcal{A})$ into the final input-output generative format, which can be represented as $(\mathcal{X},\mathcal{A})$. The input $\mathcal{X}$ is constructed by concatenating the markdown format of EHR historical events with the task-specific instruction $\mathcal{I}$: 
\begin{equation}
    \mathcal{X} = \text{Concate}(\text{Markdown}(\mathcal{R}_{\le t}), \mathcal{I})
\end{equation}
Concurrently, the target output $\mathcal{A}$ is formed by joining all items from the corresponding label set, with each item separated by a newline character.


\paragraph{Large-scale Domain Adaptation. } The initial step in our training pipeline is to adapt the base LLM to the clinical domain of EHR data by instruction-tuning on na\"ive question-answering pairs without reasoning. 
In our experiments, we found that supervising the model on both the input $\mathcal{X}$ and the target labels $\mathcal{A}$ during this phase yields markedly better performance on downstream EHR tasks than training on $\mathcal{A}$ alone. We believe this improvement stems from the fact that $\mathcal{X}$ encodes rich distributional and temporal patterns of patient trajectories, which guide the model to learn co-occurrence and sequence dependencies more effectively. Formally, we minimize the combined loss:
\begin{equation}
    \mathcal{L}_{\text{SFT}}(\Phi_{\text{LLM}}) = -\sum_{i=1}^{|\mathcal{A}|}\log \Phi_{\text{LLM}}(\mathcal{A}_i|\mathcal{A}_{<i},\mathcal{X}) -\sum_{j=1}^{|\mathcal{X}|}\log \Phi_{\text{LLM}}(\mathcal{X}_j|\mathcal{X}_{<j})
\end{equation}

\paragraph{Reasoning Enhancement.} Following the domain adaptation phase, we conduct instruction-tuning on the reasoning samples aiming at teaching the model how to perform explicit reasoning on EHR analysis tasks. To achieve this, we adopt the reasoning data generated by our automatic synthesis pipeline (as detailed in Section~\ref{section: reasoning synthesis}).

While the input $\mathcal{X}$ remains the patient's observable events and the task instruction, the output $\mathcal{A}$ is an augmented sequence that contains both a detailed reasoning chain encapsulated within the `<think>' and `</think>' tags and the final result. The training objective for this stage remains the same as that of large-scale domain adaptation to ensure consistency and reinforce the model's ability to learn from both the input sequence and the structured reasoning output.


\paragraph{Reinforcement Learning.} 
Despite the gains achieved through instruction-tuning, the model’s reasoning capacity remains limited by the quality of the generated data. To further unlock its reasoning potential, we apply Group Reward Policy Optimization (GRPO) on top of the SFT checkpoint, with the expectation that end-to-end reinforcement learning will guide the model to self-explore improved reasoning trajectories.

GRPO frames each reasoning chain as a sequential decision process, rewarding trajectories that yield correct predictions and coherent, evidence-grounded explanations. By exploring multiple candidate chains and reinforcing the most effective ones, the model learns to prioritize high-quality inference paths. 
Specifically, we design GRPO’s reward function $R_{\text{GRPO}}(\tau)$ for the inference reasoning trajectory $\tau$ as a sum of two components: format reward and accuracy reward:

\begin{equation}
R(\tau,\hat{\mathcal{A}},\mathcal{A},\mathcal{I}) = \lambda_{\mathrm{fmt}}\cdot R_{\mathrm{fmt}}(\tau)+\lambda_{\mathrm{acc}}\cdot R_{\mathrm{acc}}(\hat{\mathcal{A}},\mathcal{A},\mathcal{I})
\end{equation}
where $\hat{\mathcal{A}}$ is the predicted entity set of LLMs, $\mathcal{A}$ is the target entity set, and $\mathcal{I}$ is the type of EHR analysis task.
The format reward $R_{\mathrm{fmt}}(\tau)$ checks that whether the model wraps its chain of thought between the tags `<think>' and `</think>' and respects the three prescribed stages: Extraction, Reasoning, and Final Result:
\begin{equation}
    R_{\mathrm{fmt}}(\tau) =
    \begin{cases}
        1, & \text{if the format of }\tau\text{ is right}  \\
        0, & \text{otherwise}
    \end{cases}
\end{equation}

The accuracy reward $R_{\mathrm{acc}}(\tau)$ uses the task-specific metric. ACC for the risk-prediction task and F1 for the decision-making task are used to score the final prediction against the ground truth:

\begin{equation}
    R_{\mathrm{acc}}(\hat{\mathcal{A}},\mathcal{A}, \mathcal{I}) =
    \begin{cases}
        \text{ACC}(\hat{\mathcal{A}},\mathcal{A}), & \text{if }\mathcal{I} \ \text{is} \ \text{decision-making Task}  \\
        \text{F1}(\hat{\mathcal{A}},\mathcal{A}), & \text{if }\mathcal{I} \ \text{is} \ \text{Risk P{rediction Task}}
    \end{cases}
\end{equation}

To guard against runaway drift and hallucination, we restrict GRPO updates to just 500 examples per task, ensuring that the model refines its reasoning without straying too far from the validated chains generated during instruction tuning. This targeted reinforcement step produces a model that not only reasons more flexibly over EHR data but also maintains the reliability and interpretability of its clinical explanations.

\paragraph{Implementation Details}
For our experiments, we adopt Qwen3-1.7B, Qwen3-8B, and Qwen2.5-72B~\cite{yang2025qwen3} as our base models. The same three stage training strategy us applied to all three models. All the training are conduct based on the code of \texttt{Transformers}\footnote{\url{https://github.com/huggingface/transformers}} and \texttt{TRL}\footnote{\url{https://github.com/huggingface/trl}} framework. For GRPO stage, The weighted hyperparameters of rewards $\lambda_{\mathrm{fmt}}$ and $\lambda_{\mathrm{acc}}$ are both set to 1 and the number of rollout is 8 per sample. More detail training hyperparameters are shown in Table~\ref{tab: hyperparam}.

\begin{table}[!t]
\centering
\caption{Training hyperparameters for three base models. In Instruction Tuning, all models use a high learning rate and large batch sizes for efficient pre-training. For Reasoning Tuning and GRPO, models train for more epochs (3 and 2 respectively) to compensate for smaller, specialized datasets. During GRPO, larger models (Qwen3-8B and Qwen2.5-72B) adopt smaller learning rates (5e-7) to ensure training stability given their extensive parameters. Finally, Qwen2.5-72B's shorter maximum sequence length (8192) across Reasoning Tuning and GRPO is a necessary compromise due to computational resource limitations. }
\label{tab: hyperparam}
\vspace{6pt}
\resizebox{.9\textwidth}{!}{%
\begin{tabular}{ccccccc}
\toprule
\textbf{Training Stage} & \textbf{Model} & \textbf{LR} & \textbf{Batch Size} & \textbf{Epoch} & \textbf{Max Seq Len} & \textbf{GPU Num} \\
\midrule
\multirow{3}{*}{Instruction Tuning} & Qwen3-1.7B & 2e-5 & 128 & 1 & 8192 & 64 \\
 & Qwen3-8B & 2e-5 & 256 & 1 & 8192 & 128 \\
 & Qwen2.5-72B & 2e-5 & 512 & 1 & 8192 & 256 \\
 \midrule
\multirow{3}{*}{Reasoning Tuning} & Qwen3-1.7B & 2e-5 & 256 & 3 & 12288 & 128 \\
 & Qwen3-8B & 2e-5 & 256 & 3 & 12288 & 128 \\
 & Qwen2.5-72B & 2e-5 & 256 & 3 & 8192 & 128 \\
 \midrule
\multirow{3}{*}{GRPO} & Qwen3-1.7B & 2e-6 & 128 & 2 & 12288 & 128 \\
 & Qwen3-8B & 5e-7 & 32 & 2 & 12288 & 128 \\
 & Qwen2.5-72B & 5e-7 & 32 & 2 & 8192 & 128 \\
 \bottomrule
\end{tabular}%
}
\end{table}

\subsection{Evaluation}
Lastly, we introduce the evaluation settings, including used benchmarks, considered baselines, and metrics.

\subsubsection{Benchmarks}
\label{sec:benchmarks}

In this section, we present the benchmarks considered in this paper and describe how we reformatted them to align with our generative evaluation pipeline, accompanied by detailed case demonstrations.



\paragraph{\benchname.} In \benchname, we've included two categories of tasks. One of these, decision-making tasks, requires the model to predict multiple possible target answers. For evaluation, a prediction is only considered correct if the model's predicted medical entities precisely match those in the target answer. However, since most baseline models aren't fine-tuned on MIMIC-IV data, we generate a candidate set by randomly sampling from the task's target label set. Baseline models then simply select medical entities from this candidate set. It's worth noting that our proposed model, \modelname, can directly generate answers without needing this candidate set, demonstrating its practical usability. To ensure the difficulty for baseline models selecting from options is comparable to \modelname directly generating answers, we set the candidate set size for each sample to 100 (including the number of target labels). If a task has fewer than 100 target label types, we provide all available target labels for the model to choose from. The example is shown in Case~\ref{case: mimic-bench}.

\paragraph{MIMIC-IV-CDM Benchmark.} MIMIC-IV-CDM meticulously selected four diseases—appendicitis, cholecystitis, diverticulitis, and pancreatitis—to test model diagnostic accuracy. Unlike \benchname and EHRSHOT, the historical event information in MIMIC-IV-CDM is not arranged chronologically. Instead, it extracts the most recent information for each event to serve as context. Consequently, the context in MIMIC-IV-CDM is less noisy, and the task is simpler. To maintain consistency with the other two benchmarks, we also reorganized the MIMIC-IV-CDM data into a Markdown-formatted free text. Given that the information within it lacks timestamps, the title for each event only includes the event name without the time it occurred. The example case of the free-text input is shown in Case~\ref{case: mimic-iv-cdm}.

\paragraph{EHRSHOT Benchmark.}
EHRSHOT, a public dataset collected by Stanford University, comprises electronic health records from 7,000 patients. Its publicly available portion includes 14 risk-prediction tasks categorized into three subtypes: Operational Outcomes, Anticipating Lab Test Results, and Assignment of New Diagnoses. EHRSHOT is originally designed for traditional machine learning models, presents a challenge for Large Language Models (LLMs) due to its direct use of various medical codes, which are inherently difficult for LLMs to interpret. To address this, we've reformatted the EHRSHOT data into a Markdown-like free-text, similar to \benchname, enabling direct testing with LLMs. Specifically, we utilized the descriptive mapping lexicon provided within EHRSHOT to convert item codes into natural language text. However, certain codes, such as CPT4 and ICD10PCS, could not be directly transformed. Following previous work on evaluating LLMs with EHRSHOT~\cite{hegselmann2025large}, we manually added extra mapping files to resolve these inconsistencies\footnote{CPT4 mapping: \url{ https://gist.github.com/lieldulev/439793dc3c5a6613b661c33d71fdd185}\\ICD10PCS mapping: \url{https://hcup-us.ahrq.gov/toolssoftware/ccsr/PRCCSR_v2025-1.zip}}. Furthermore, for EHRSHOT's measurement and observation events, we semantically clustered the 24 most frequently occurring items. We then enriched these items with additional information, including their units, normal value ranges, and anomaly indicators, significantly enhancing the informational content of EHRSHOT. The example case of the free-text input is shown in Case~\ref{case: ehrshot}.

\subsubsection{Metrics}
\label{section: metric}
Based on the two high-level EHR analysis task types, \emph{i.e.}, decision-making and risk-prediction, different evaluation metrics are adopted to quantitatively reflect the performance of LLMs, as formulated below.

\paragraph{Decision-Making Tasks.}
For each decision-making sample, let the ground-truth target set be $\mathcal{A}$ and the model’s predicted set be $\hat{\mathcal{A}}$. We adopt the precision, recall, F1 score to evaluate prediction accuracy in this multi-label setting:
\begin{equation}
    \text{Precision}=\frac{|\mathcal{A}\cap \hat{\mathcal{A}}|}{|\hat{\mathcal{A}}|},\quad \text{Recall}=\frac{|\mathcal{A}\cap \hat{\mathcal{A}}|}{|\mathcal{A}|},\quad
    \text{F1} = \frac{2 \times \text{Precision} \times \text{Recall}}{\text{Precision} + \text{Recall}}.
\end{equation}

\paragraph{Risk-Prediction Tasks.}
For risk-prediction tasks, we follow prior work~\cite{NEURIPS2023_d42db1f7} and adopt the area under the receiver operating characteristic curve (AUROC) as the evaluation metric. AUROC measures a model’s ability to distinguish between positive and negative outcomes across all decision thresholds, and can be written as
\begin{equation}
\mathrm{AUROC} = \int_{0}^{1} \mathrm{TPR}(t)\, d\mathrm{FPR}(t),
\end{equation}
where $\mathrm{TPR}$ and $\mathrm{FPR}$ denote the true positive and false positive rates, respectively.


\subsubsection{Baselines}
\label{sec:baselines}
The following LLMs are used as baselines for comparison: 

\paragraph{Qwen2.5/3 Series.} Developed by Alibaba Cloud, the Qwen2.5/3 series~\cite{yang2025qwen3} offers versatile, general-purpose multilingual models with a unique `thinking' and `non-thinking' hybrid reasoning engine and Mixture-of-Experts (MoE) architecture for efficient inference. Trained on 36 trillion tokens across over 100 languages, Qwen3 excels in general reasoning, coding, and agent capabilities. While not specifically medical, its robust multilingual support and advanced reasoning could serve as a powerful general foundation for processing diverse medical literature or assisting in global health information dissemination.
\paragraph{Medgemma 4B/27B.} Created by Google, MedGemma~\cite{sellergren2025medgemma} is a collection of Gemma 3 variants specifically optimized for medical text and image comprehension. Its key innovation lies in its multimodal capability, seamlessly integrating natural language processing with computer vision to analyze medical images (e.g., X-rays, histopathology) alongside textual patient data. Trained on extensive de-identified medical datasets, MedGemma demonstrates superior performance on clinical benchmarks like MedQA and is designed to accelerate healthcare AI applications while supporting local deployment for data privacy.

\paragraph{Llama3 Series.} The Llama3 Series~\cite{DBLP:journals/corr/abs-2407-21783}, developed by Meta, comprises a family of powerful open-source general-purpose LLMs ranging from 8 billion to 405 billion parameters, with evolving multimodal capabilities in Llama 3.2. Trained on over 15 trillion tokens, these models excel in broad applications, including complex reasoning, coding, and multilingual understanding. While not inherently specialized for medicine, the Llama3 architecture serves as a robust foundation, with fine-tuned variants like Me-LLaMA demonstrating superior performance in various medical question-answering and summarization tasks, highlighting its adaptability for domain-specific healthcare solutions.

\paragraph{OpenBioLLM.} Developed by Saama AI Labs, OpenBioLLM~\cite{dorfner2024biomedical} is an 8-billion-parameter open-source language model specifically designed for the biomedical field, built upon the Llama 3 architecture. It is meticulously fine-tuned on high-quality biomedical data, enabling it to accurately understand and generate text for tasks like clinical entity recognition, medical question answering, and patient data de-identification. OpenBioLLM consistently outperforms larger general-purpose models on diverse biomedical benchmarks, making it a highly effective and transparent choice for healthcare and life sciences applications.

\paragraph{Baichuan M2 32B.} Baichuan-M2~\cite{wang2025baichuan}, from Baichuan Intelligent, stands out as the first open-source LLM specifically optimized for medical scenarios, trained "from scratch" on an unprecedented 20 trillion tokens of high-quality medical and general data. This dedicated training, coupled with a multi-stage curriculum learning approach, enables it to achieve deep medical expertise, outperforming models five times its size in clinical practice benchmarks while maintaining strong general capabilities. Its specialized focus makes it highly relevant for tasks such as clinical decision support, medical education, and rare disease diagnosis.
\paragraph{GPT-4o.} OpenAI's GPT-4o (omni)~\cite{hurst2024gpt} is a flagship multimodal model designed for natural human-computer interaction, accepting and generating any combination of text, audio, image, and video through a single, end-to-end neural network . It offers real-time responsiveness and enhanced tokenization for non-English languages, making it faster and more cost-effective. While a general-purpose model, its advanced multimodal understanding and reasoning capabilities could be applied to complex medical cases involving diverse data types, from patient consultations (audio/text) to diagnostic images.
\paragraph{DeepSeek R1.} Developed by DeepSeek, R1~\cite{DBLP:journals/corr/abs-2501-12948} is an open-source AI model released in January 2025, primarily focused on pushing the limits of reinforcement learning as a post-training technique for complex reasoning. Utilizing a Mixture-of-Experts (MoE) architecture, it achieves high accuracy in mathematics, programming, and general reasoning tasks at a significantly lower operating cost. Notably, DeepSeek R1 has shown strong performance on medical benchmarks like MedQA~\cite{jin2021disease}, demonstrating its potential for efficient and accurate reasoning in healthcare contexts.
\paragraph{GPT-OSS.} The GPT-OSS (Open-Source Series)~\cite{gptoss} is OpenAI's first open-source GPT-class model since GPT-2, released under an Apache 2.0 license to provide powerful reasoning and agentic capabilities to the open community. The models, gpt-oss-120b and gpt-oss-20b, utilize a Mixture-of-Experts (MoE) architecture for efficient inference and support a long 128k token context window. While a general-purpose reasoning model, its performance on complex, health-related queries is noteworthy; the gpt-oss-120b model achieved a score of 30\% on the HealthBench benchmark, outperforming OpenAI's own o4-mini and o3-mini models on this specific task. This demonstrates its potential as a strong, customizable foundation for a variety of applications, including those with intricate medical reasoning requirements.

\section{Data Availability}
The data source for \benchname and \traindataname derives from MIMIC-IV, a public EHR resource. Both datasets are under review of PhysioNet\footnote{\url{https://physionet.org/}}.

\section{Code Availability}
All source codes of this paper have been released in \url{https://github.com/MAGIC-AI4Med/EHR-R1}.


\newpage
\bibliographystyle{unsrt} 
\bibliography{references} 

\section{Acknowledgments}
This work is supported by the National Key R\&D Program of China (No. 2022ZD0160702), and the Scientific Research Innovation Capability Support Project for Young Faculty~(ZY-GXQNJSKYCXNLZCXM-I22).

\section{Author Contributions}
All listed authors clearly meet the ICMJE 4 criteria. Y.L., C.W., and J.L. contribute equally to this work. Y.W. and W.X. are the corresponding authors. Specifically, Y.L., C.W., J.L., S.J., P.Q., H.W., Y.Y., S.Z., J.W., Q.F., J.G., Y.Z., Y.W., Y.W., and W.X. all make contributions to the conception or design of the work, and Y.L., C.W., J.L. further perform acquisition, analysis, or interpretation of data for the work. In writing,  Y.L., C.W., J.L. draft the work. S.J., P.Q., H.W., Y.Y., S.Z., J.W., Q.F., J.G., Y.Z., Y.W., Y.W., and W.X. review it critically for important intellectual content. All authors approve of the version to be published and agree to be accountable for all aspects of the work to ensure that questions related to the accuracy or integrity of any part of the work are appropriately investigated and resolved.

\clearpage
\section{Supplementary}
\setcounter{table}{0}   
\setcounter{figure}{0}
\renewcommand{\tablename}{Supplementary Table}
\renewcommand{\figurename}{Supplementary Figure}

\subsection{Details of Experiments}

\begin{table}[!h]
\centering
\caption{\textbf{Accurate results of decision making tasks in \benchname.} The performance are measured with the metric F1 score and shown in in the format `mean $\pm$ std'. }
\label{tab: deicsion making}
\resizebox{\textwidth}{!}{%
\begin{tabular}{lcccccccccc}
\toprule
\textbf{Tasks}     & \multicolumn{1}{c}{\textbf{Llama3.3-70B}} & \multicolumn{1}{c}{\textbf{GPT-4o}} & \multicolumn{1}{c}{\textbf{Qwen2.5-72B}} & \multicolumn{1}{c}{\textbf{OpenBioLLM-70B}} & \multicolumn{1}{c}{\textbf{Baichuan-M2-32B}} & \multicolumn{1}{c}{\textbf{Medgemma-27B}} & \multicolumn{1}{c}{\textbf{GPT-OSS-120B}} & \multicolumn{1}{c}{\textbf{DeepSeek-R1}} & \multicolumn{1}{c}{\textbf{Qwen3-235B}} & \multicolumn{1}{c}{\textbf{EHR-R1-72B}} \\
\midrule
Admissions         & 0.1438$\pm$0.0169                         & 0.3151$\pm$0.0208                   & 0.1452$\pm$0.0172                        & 0.0787$\pm$0.0122                           & 0.0633$\pm$0.0074                            & 0.2172$\pm$0.0164                         & 0.1029$\pm$0.0141                         & 0.1728$\pm$0.0166                        & 0.0974$\pm$0.0134                       & 0.7512$\pm$0.0186                       \\
OMR                & 0.0569$\pm$0.0059                         & 0.0582$\pm$0.0080                   & 0.1591$\pm$0.0092                        & 0.0598$\pm$0.0077                           & 0.1373$\pm$0.0088                            & 0.1739$\pm$0.0099                         & 0.0835$\pm$0.0092                         & 0.0115$\pm$0.0032                        & 0.1989$\pm$0.0107                       & 0.9242$\pm$0.0059                       \\
EMAR               & 0.2941$\pm$0.0156                         & 0.5543$\pm$0.0213                   & 0.5087$\pm$0.0227                        & 0.2558$\pm$0.0166                           & 0.2639$\pm$0.0114                            & 0.3858$\pm$0.0183                         & 0.4118$\pm$0.0198                         & 0.5290$\pm$0.0225                        & 0.5801$\pm$0.0179                       & 0.6503$\pm$0.0192                       \\
Procedures ICD     & 0.1737$\pm$0.0085                         & 0.1693$\pm$0.0138                   & 0.2049$\pm$0.0156                        & 0.0848$\pm$0.0118                           & 0.1272$\pm$0.0073                            & 0.2001$\pm$0.0103                         & 0.2074$\pm$0.0172                         & 0.2770$\pm$0.0173                        & 0.1976$\pm$0.0139                       & 0.4079$\pm$0.0189                       \\
Procedures CCS     & 0.1634$\pm$0.0109                         & 0.2718$\pm$0.0196                   & 0.2557$\pm$0.0158                        & 0.0689$\pm$0.0106                           & 0.1466$\pm$0.0086                            & 0.2047$\pm$0.0116                         & 0.2971$\pm$0.0186                         & 0.3064$\pm$0.0174                        & 0.2907$\pm$0.0160                       & 0.5340$\pm$0.0189                       \\
Diagnoses ICD      & 0.2432$\pm$0.0088                         & 0.1767$\pm$0.0116                   & 0.2805$\pm$0.0129                        & 0.0786$\pm$0.0074                           & 0.1880$\pm$0.0067                            & 0.2681$\pm$0.0088                         & 0.0972$\pm$0.0078                         & 0.2058$\pm$0.0119                        & 0.2470$\pm$0.0111                       & 0.4088$\pm$0.0123                       \\
Diagnoses CCS      & 0.3972$\pm$0.0107                         & 0.2847$\pm$0.0121                   & 0.4529$\pm$0.0122                        & 0.1880$\pm$0.0119                           & 0.3366$\pm$0.0080                            & 0.3760$\pm$0.0099                         & 0.2293$\pm$0.0137                         & 0.3451$\pm$0.0131                        & 0.3992$\pm$0.0123                       & 0.5654$\pm$0.0134                       \\
Labevents          & 0.4558$\pm$0.0152                         & 0.1667$\pm$0.0108                   & 0.4788$\pm$0.0123                        & 0.0702$\pm$0.0070                           & 0.4001$\pm$0.0126                            & 0.4901$\pm$0.0138                         & 0.1878$\pm$0.0096                         & 0.2625$\pm$0.0107                        & 0.4740$\pm$0.0135                       & 0.6681$\pm$0.0178                       \\
Microbiologyevents & 0.1915$\pm$0.0091                         & 0.2437$\pm$0.0183                   & 0.2826$\pm$0.0174                        & 0.0458$\pm$0.0093                           & 0.1494$\pm$0.0078                            & 0.3325$\pm$0.0159                         & 0.2250$\pm$0.0200                         & 0.2627$\pm$0.0158                        & 0.2888$\pm$0.0140                       & 0.7101$\pm$0.0215                       \\
Services           & 0.2055$\pm$0.0185                         & 0.2421$\pm$0.0188                   & 0.1888$\pm$0.0150                        & 0.0902$\pm$0.0129                           & 0.0860$\pm$0.0083                            & 0.2060$\pm$0.0174                         & 0.2599$\pm$0.0199                         & 0.2834$\pm$0.0202                        & 0.1938$\pm$0.0191                       & 0.8451$\pm$0.0169                       \\
Transfers          & 0.2997$\pm$0.0182                         & 0.3686$\pm$0.0217                   & 0.3697$\pm$0.0227                        & 0.2668$\pm$0.0203                           & 0.1831$\pm$0.0092                            & 0.3626$\pm$0.0221                         & 0.3035$\pm$0.0211                         & 0.3500$\pm$0.0219                        & 0.3968$\pm$0.0195                       & 0.7634$\pm$0.0153                       \\
POE                & 0.2276$\pm$0.0101                         & 0.1716$\pm$0.0147                   & 0.2401$\pm$0.0121                        & 0.1793$\pm$0.0144                           & 0.1457$\pm$0.0083                            & 0.1880$\pm$0.0091                         & 0.1220$\pm$0.0139                         & 0.1699$\pm$0.0126                        & 0.2280$\pm$0.0123                       & 0.5482$\pm$0.0200                       \\
Radiology          & 0.2115$\pm$0.0136                         & 0.2943$\pm$0.0197                   & 0.2418$\pm$0.0192                        & 0.1364$\pm$0.0164                           & 0.1408$\pm$0.0090                            & 0.2583$\pm$0.0169                         & 0.2540$\pm$0.0198                         & 0.2982$\pm$0.0205                        & 0.2383$\pm$0.0163                       & 0.5542$\pm$0.0210                       \\
Prescriptions      & 0.2512$\pm$0.0110                         & 0.4533$\pm$0.0218                   & 0.3390$\pm$0.0141                        & 0.2772$\pm$0.0192                           & 0.2124$\pm$0.0086                            & 0.2705$\pm$0.0124                         & 0.2271$\pm$0.0158                         & 0.3157$\pm$0.0194                        & 0.3761$\pm$0.0162                       & 0.8560$\pm$0.0116                       \\
Prescriptions ATC  & 0.2375$\pm$0.0128                         & 0.2686$\pm$0.0183                   & 0.3996$\pm$0.0183                        & 0.1443$\pm$0.0162                           & 0.1848$\pm$0.0091                            & 0.2739$\pm$0.0137                         & 0.1617$\pm$0.0149                         & 0.2316$\pm$0.0174                        & 0.2974$\pm$0.0134                       & 0.7891$\pm$0.0138                       \\
Medrecon           & 0.2551$\pm$0.0114                         & 0.3210$\pm$0.0159                   & 0.3030$\pm$0.0133                        & 0.0171$\pm$0.0048                           & 0.2343$\pm$0.0102                            & 0.2436$\pm$0.0113                         & 0.1139$\pm$0.0102                         & 0.2961$\pm$0.0149                        & 0.2545$\pm$0.0133                       & 0.3319$\pm$0.0131                       \\
Medrecon ATC       & 0.2524$\pm$0.0099                         & 0.3383$\pm$0.0137                   & 0.4006$\pm$0.0128                        & 0.0271$\pm$0.0049                           & 0.2647$\pm$0.0093                            & 0.3345$\pm$0.0133                         & 0.1002$\pm$0.0100                         & 0.2842$\pm$0.0159                        & 0.4003$\pm$0.0128                       & 0.3981$\pm$0.0142                       \\
ED Diagnoses ICD   & 0.1740$\pm$0.0076                         & 0.2583$\pm$0.0150                   & 0.2320$\pm$0.0131                        & 0.0691$\pm$0.0087                           & 0.1421$\pm$0.0064                            & 0.1560$\pm$0.0076                         & 0.2231$\pm$0.0144                         & 0.2514$\pm$0.0154                        & 0.2310$\pm$0.0110                       & 0.4319$\pm$0.0168                       \\
ED Diagnoses CCS   & 0.2593$\pm$0.0098                         & 0.3877$\pm$0.0173                   & 0.3595$\pm$0.0127                        & 0.1365$\pm$0.0127                           & 0.2271$\pm$0.0079                            & 0.2321$\pm$0.0076                         & 0.4240$\pm$0.0151                         & 0.3982$\pm$0.0153                        & 0.3669$\pm$0.0118                       & 0.5674$\pm$0.0149                       \\
Ingredientevents   & 0.6613$\pm$0.0124                         & 0.5826$\pm$0.0146                   & 0.7537$\pm$0.0101                        & 0.4662$\pm$0.0117                           & 0.5176$\pm$0.0077                            & 0.7131$\pm$0.0119                         & 0.4914$\pm$0.0165                         & 0.5704$\pm$0.0166                        & 0.7489$\pm$0.0110                       & 0.8739$\pm$0.0080                       \\
Datetimeevents     & 0.3659$\pm$0.0164                         & 0.3438$\pm$0.0196                   & 0.5096$\pm$0.0206                        & 0.2956$\pm$0.0192                           & 0.1931$\pm$0.0099                            & 0.4280$\pm$0.0181                         & 0.0572$\pm$0.0082                         & 0.1557$\pm$0.0131                        & 0.4089$\pm$0.0167                       & 0.9758$\pm$0.0043                       \\
Procedureevents    & 0.0745$\pm$0.0077                         & 0.0938$\pm$0.0128                   & 0.1030$\pm$0.0108                        & 0.1231$\pm$0.0146                           & 0.0655$\pm$0.0059                            & 0.1628$\pm$0.0150                         & 0.0787$\pm$0.0122                         & 0.0884$\pm$0.0115                        & 0.1192$\pm$0.0108                       & 0.8377$\pm$0.0179                       \\
Inputevents        & 0.3504$\pm$0.0158                         & 0.3979$\pm$0.0204                   & 0.4516$\pm$0.0192                        & 0.1565$\pm$0.0163                           & 0.2457$\pm$0.0085                            & 0.4030$\pm$0.0181                         & 0.2621$\pm$0.0159                         & 0.2705$\pm$0.0193                        & 0.3911$\pm$0.0152                       & 0.8763$\pm$0.0129                       \\
Outputevents       & 0.6730$\pm$0.0180                         & 0.8107$\pm$0.0170                   & 0.8231$\pm$0.0154                        & 0.5745$\pm$0.0200                           & 0.4049$\pm$0.0087                            & 0.6949$\pm$0.0184                         & 0.7556$\pm$0.0190                         & 0.8012$\pm$0.0176                        & 0.7794$\pm$0.0168                       & 0.9176$\pm$0.0116                       \\
\midrule
Average            & 0.2758$\pm$0.0025                         & 0.3155$\pm$0.0037                   & 0.3535$\pm$0.0031                        & 0.1621$\pm$0.0027                           & 0.2108$\pm$0.0018                            & 0.3157$\pm$0.0027                         & 0.2365$\pm$0.0035                         & 0.2974$\pm$0.0033                        & 0.3418$\pm$0.0025                       & 0.6744$\pm$0.0029   \\
\bottomrule
\end{tabular}%
}
\end{table}

\begin{table}[!h]
\centering
\caption{\textbf{Accurate results of ablation experiments.} The table presents the performance of different model configurations across various decision-making tasks. We evaluate our method on base models of different sizes, including Qwen3-1.7B, Qwen3-8B, Qwen2.5-72B, and Qwen3-235B. The performances are measured with the metric F1 score and shown in the format `mean $\pm$ std', with the average performance across all tasks provided in the final column.}
\label{tab: ablation}
\resizebox{\textwidth}{!}{%
\begin{tabular}{lcccccccc}
\toprule
\multicolumn{1}{c}{}                                      & \multicolumn{7}{c}{\textbf{Decision Making Tasks}}                                                                                                                                                                                   &                                    \\
\cmidrule{2-8}
\multicolumn{1}{c}{\multirow{-2}{*}{\textbf{Methods}}}    & \textbf{Transfer}              & \textbf{Service}               & \textbf{Procedure}             & \textbf{Test\&Exam}          & \textbf{Diagnose}              & \textbf{Treatment}             & \textbf{ICU Event}             & \multirow{-2}{*}{\textbf{Average}} \\
\midrule
\multicolumn{9}{c}{\cellcolor[HTML]{F2F2F2}\textbf{Qwen3-1.7B}}    \\
\midrule
\textbf{Base Model}                                       & 0.1332$\pm$0.0087 & 0.1166$\pm$0.0059 & 0.0820$\pm$0.0071 & 0.1292$\pm$0.0053 & 0.1344$\pm$0.0035 & 0.1225$\pm$0.0032 & 0.3119$\pm$0.0064 & 0.1544$\pm$0.0020     \\
\textbf{Base Model (w/ reasoning   inference)}            & 0.1530$\pm$0.0107 & 0.0996$\pm$0.0093 & 0.1068$\pm$0.0087 & 0.0993$\pm$0.0055 & 0.1493$\pm$0.0055 & 0.1182$\pm$0.0051 & 0.2374$\pm$0.0066 & 0.1393$\pm$0.0024     \\
\textbf{EHR-R1 (w/o reasoning   training)}                & 0.6510$\pm$0.0144 & 0.4953$\pm$0.0127 & 0.3037$\pm$0.0119 & 0.5251$\pm$0.0081 & 0.3319$\pm$0.0071 & 0.4756$\pm$0.0078 & 0.7773$\pm$0.0071 & 0.5060$\pm$0.0034     \\
\textbf{EHR-R1 (full training)}                           & 0.6257$\pm$0.0153 & 0.6106$\pm$0.0134 & 0.2663$\pm$0.0121 & 0.6135$\pm$0.0089 & 0.3302$\pm$0.0066 & 0.4916$\pm$0.0067 & 0.8031$\pm$0.0072 & 0.5300$\pm$0.0029     \\
\textbf{EHR-R1 (full training, w/   reasoning inference)} & 0.6863$\pm$0.0148 & 0.6232$\pm$0.0150 & 0.2924$\pm$0.0136 & 0.6350$\pm$0.0089 & 0.3610$\pm$0.0075 & 0.5035$\pm$0.0075 & 0.8280$\pm$0.0060 & 0.5438$\pm$0.0035     \\
\midrule
\multicolumn{9}{c}{\cellcolor[HTML]{F2F2F2}\textbf{Qwen3-8B}}  \\
\midrule
\textbf{Base Model}                                       & 0.1585$\pm$0.0119 & 0.1445$\pm$0.0080 & 0.1665$\pm$0.0098 & 0.2082$\pm$0.0062 & 0.2478$\pm$0.0048 & 0.2411$\pm$0.0062 & 0.4358$\pm$0.0077 & 0.2425$\pm$0.0029     \\
\textbf{Base Model (w/ reasoning   inference)}            & 0.2370$\pm$0.0123 & 0.1815$\pm$0.0091 & 0.2083$\pm$0.0083 & 0.1639$\pm$0.0062 & 0.2890$\pm$0.0056 & 0.1903$\pm$0.0062 & 0.3527$\pm$0.0061 & 0.2286$\pm$0.0026     \\
\textbf{EHR-R1 (w/o reasoning   training)}                & 0.6670$\pm$0.0144 & 0.5372$\pm$0.0145 & 0.3617$\pm$0.0132 & 0.5793$\pm$0.0078 & 0.3917$\pm$0.0070 & 0.5155$\pm$0.0073 & 0.8249$\pm$0.0069 & 0.5549$\pm$0.0038     \\
\textbf{EHR-R1 (full training)}                           & 0.6681$\pm$0.0145 & 0.5661$\pm$0.0117 & 0.3192$\pm$0.0122 & 0.5681$\pm$0.0088 & 0.3877$\pm$0.0071 & 0.5348$\pm$0.0070 & 0.8462$\pm$0.0057 & 0.5616$\pm$0.0031     \\
\textbf{EHR-R1 (full training, w/   reasoning inference)} & 0.7091$\pm$0.0138 & 0.6558$\pm$0.0124 & 0.3531$\pm$0.0135 & 0.6764$\pm$0.0092 & 0.4062$\pm$0.0072 & 0.5523$\pm$0.0071 & 0.8521$\pm$0.0063 & 0.5894$\pm$0.0037     \\
\midrule
\multicolumn{9}{c}{\cellcolor[HTML]{F2F2F2}\textbf{Qwen2.5-72B}}  \\
\midrule
\textbf{Base Model}                                       & 0.2564$\pm$0.0142 & 0.2146$\pm$0.0092 & 0.2283$\pm$0.0116 & 0.2888$\pm$0.0074 & 0.3310$\pm$0.0053 & 0.3900$\pm$0.0074 & 0.5289$\pm$0.0072 & 0.3323$\pm$0.0031     \\
\textbf{EHR-R1 (w/o reasoning   training)}                & 0.7049$\pm$0.0151 & 0.5572$\pm$0.0131 & 0.4368$\pm$0.0133 & 0.6412$\pm$0.0085 & 0.4304$\pm$0.0077 & 0.5619$\pm$0.0072 & 0.8808$\pm$0.0061 & 0.6039$\pm$0.0037   \\
\textbf{EHR-R1 (full training)}                           & 0.7176$\pm$0.0155 & 0.6623$\pm$0.0140 & 0.4322$\pm$0.0126 & 0.6797$\pm$0.0085 & 0.4506$\pm$0.0071 & 0.5796$\pm$0.0070 & 0.8840$\pm$0.0060 & 0.6281$\pm$0.0034     \\
\textbf{EHR-R1 (full training, w/   reasoning inference)} & 0.7379$\pm$0.0143 & 0.6905$\pm$0.0113 & 0.4489$\pm$0.0143 & 0.7013$\pm$0.0092 & 0.4702$\pm$0.0070 & 0.5918$\pm$0.0060 & 0.8894$\pm$0.0054 & 0.6418$\pm$0.0031     \\
\bottomrule
\end{tabular}%
}
\end{table}
\newpage

\subsection{Details of \benchname}

\begin{table}[!h]
\centering
\caption{Details of task information in \benchname. This table presents a comprehensive overview of the 42 tasks, categorized into 2 main Task Types and further divided into 12 SubTypes. The Target Event column specifies the type of event associated with the target label for each task. For decision making tasks, the Item Name indicates the column name in MIMIC-IV where the target label is located. It's important to note that Risk Prediction tasks do not have an Item Name as their target labels are not specific entities. Columns marked with `*' represent additional columns created through manual mapping. The Candidates column shows the size of the target candidate set for each task.}
\label{tab: task info}
\vspace{6pt}
\resizebox{\linewidth}{!}{%
\begin{tabular}{ccccccc}

\toprule
\textbf{Task Type} & \textbf{SubType} & \textbf{Task} & \textbf{Target Event} & \textbf{Item Name} & \textbf{Metric} & \textbf{Candidates} \\
\toprule
\multirow{24}{*}{Decision Making} & \multirow{2}{*}{Reassignment} & Admissions & \textit{admissions} & \textit{admission\_type} & F1 & 8 \\
 &  & Transfer & \textit{transfer} & \textit{eventtype} & F1 & 39 \\
\cmidrule{2-7}
 & \multirow{4}{*}{Test \& Exam} & OMR & \textit{omr} & \textit{result\_name} & F1 & 11 \\
 &  & Labevents & \textit{labevents} & \textit{item\_name} & F1 & 698 \\
 &  & Microbiologyevents & \textit{microbiologyevents} & \textit{test\_name} & F1 & 165 \\
 &  & Radiology & \textit{radiology} & \textit{exam\_name} & F1 & 961 \\
\cmidrule{2-7}
 & \multirow{4}{*}{Diagnoses} & Diagnose ICD & \textit{diagnose\_icd} & \textit{diagnoses} & F1 & 24467 \\
 &  & Diagnose CCS & \textit{diagnose\_icd} & \textit{CCT Type}* & F1 & 279 \\
 &  & Diagnosis ICD & \textit{diagnosis\_icd} & \textit{icd\_title} & F1 & 13171 \\
 &  & Diagnosis CCS & \textit{diagnosis\_icd} & \textit{CCS Type}* & F1 & 271 \\
\cmidrule{2-7}
 & \multirow{2}{*}{Procedures} & Procedures ICD & \textit{procedures\_icd} & \textit{procedures} & F1 & 11098 \\
 &  & Procedures CCS & \textit{procedures\_icd} & \textit{CCS Type}* & F1 & 230 \\
\cmidrule{2-7}
 & \multirow{2}{*}{Services} & Services & \textit{services} & \textit{curr\_service} & F1 & 18 \\
 &  & POE & \textit{poe} & \textit{order\_type} & F1 & 15 \\
\cmidrule{2-7}
 & \multirow{5}{*}{Treatments} & EMAR & \textit{emr} & \textit{medication} & F1 & 4153 \\
 &  & Prescriptions & \textit{prescriptions} & \textit{drug} & F1 & 9233 \\
 &  & Prescriptions ATC & \textit{prescriptions} & \textit{ATC Type}* & F1 & 913 \\
 &  & Medrecon & \textit{medrecon} & \textit{name} & F1 & 18641 \\
 &  & Medrecon ATC & \textit{medrecon} & \textit{ATC Type}* & F1 & 899 \\
  \cmidrule{2-7}
 & \multirow{5}{*}{ICU Event} & Ingredientevents & \textit{ingredientevents} & \textit{item\_name} & F1 & 15 \\
 &  & Datetimeevents & \textit{datetimeevents} & \textit{item\_name} & F1 & 137 \\
 &  & Procedureevents & \textit{procedureevents} & \textit{item\_name} & F1 & 138 \\
 &  & Inputevents & \textit{inputevents} & \textit{item\_name} & F1 & 222 \\
 &  & Outputevents & \textit{outputevents} & \textit{item\_name} & F1 & 63 \\
\toprule
\multirow{18}{*}{Risk Prediction} & \multirow{2}{*}{Transfer} & ED Hospitalization & \textit{edstays} & \textit{-} & AUROC & 2 \\
 &  & ED ICU Tranfer 12hour & \textit{edstays} & \textit{-} & AUROC & 2 \\
 \cmidrule{2-7}
 & Critical Outcomes & ED Critical Outcomes & \textit{edstays} & \textit{-} & AUROC & 2 \\
 \cmidrule{2-7}
 & \multirow{4}{*}{Readmission} & Readmission 30day & \textit{discharge} & \textit{-} & AUROC & 2 \\
 &  & Readmission 60day & \textit{discharge} & \textit{-} & AUROC & 2 \\
 &  & ICU Readmission & \textit{icustays} & \textit{-} & AUROC & 2 \\
 &  & ED Reattendance 3day & \textit{edstays} & \textit{-} & AUROC & 2 \\
 \cmidrule{2-7}
 & \multirow{4}{*}{LengthOfStay} & LengthOfStay 3day & \textit{admissions} & \textit{-} & AUROC & 2 \\
 &  & LengthOfStay 7day & \textit{admissions} & \textit{-} & AUROC & 2 \\
 &  & ICU Stay 7day & \textit{icustays} & \textit{-} & AUROC & 2 \\
 &  & ICU Stay 14day & \textit{icustays} & \textit{-} & AUROC & 2 \\
 \cmidrule{2-7}
 & \multirow{7}{*}{Mortality} & ED Inpatient Mortality & \textit{edstays} & \textit{-} & AUROC & 2 \\
 &  & Inpatient Mortality & \textit{admissions} & \textit{-} & AUROC & 2 \\
 &  & ICU Mortality 1day & \textit{icustays} & \textit{-} & AUROC & 2 \\
 &  & ICU Mortality 2day & \textit{icustays} & \textit{-} & AUROC & 2 \\
 &  & ICU Mortality 3day & \textit{icustays} & \textit{-} & AUROC & 2 \\
 &  & ICU Mortality 7day & \textit{icustays} & \textit{-} & AUROC & 2 \\
 &  & ICU Mortality 14day & \textit{icustays} & \textit{-} & AUROC & 2 \\
\bottomrule

\end{tabular}%
}
\end{table}

\newpage
\subsection{Prompt Collection}
\begin{prompt}
\label{prompt: evaluation}
\textbf{Reasoning Chain Evaluation Instruction for Clinicians} \newline
Task Description: Evaluation of Clinical Reasoning Chain Validity\newline
I. Background\newline
We have developed a large language model (LLM) designed to extract key information from electronic health records (EHRs) and perform clinical reasoning based on that information. For evaluation purposes, we have hidden the full EHR context, presenting only the critical information extracted by the model and the final diagnostic conclusion. Your task is to assess the relationship and logical consistency between the extracted information and the final conclusion.\newline
\newline
II. Data Description\newline
The model's response consists of three main sections: \#\#Extraction, \#\#Reasoning, and \#\#Final Result. You need to evaluate whether the content in \#\#Extraction and \#\#Reasoning adequately supports the \#\#Final Result and provide your final score in the Score column.\newline
\newline
III. Evaluation Objective\newline
Using the provided extracted information, you must determine if it is sufficient and accurate to support the model's final diagnostic conclusion. Specifically, you need to confirm:\newline
 - Whether the key information extracted by the model is relevant to the diagnosis and accurately expressed.\newline
 - Whether the final diagnostic conclusion is fully supported by the extracted information, without any missing essential evidence.\newline
\newline
Note: The model's reasoning result is considered correct and does not need to be validated for correctness.\newline
\newline
IV. Evaluation Process\newline
Please read the entire reasoning chain, including the extracted information and the diagnostic conclusion. Based on your professional judgment, evaluate the degree of correspondence between the information and the conclusion. The scoring criteria are as follows (five levels):\newline
 - \textbf{5 points (Very Satisfied)}: The extracted information is mostly accurate and largely supports the conclusion.\newline
 - \textbf{4 points (Satisfied)}: The extracted information supports most of the conclusion. A small part of the conclusion lacks evidence but is reasonably justified.\newline
 - \textbf{3 points (Average)}: The extracted information supports only a small part of the conclusion. The final result contains a large portion of content that is not adequately supported.\newline
 - \textbf{2 points (Poor)}: The extracted information is related to the final result, but it fails to support the conclusion.\newline
 - \textbf{1 point (Very Poor)}: The extracted information is highly irrelevant to the diagnostic conclusion. The conclusion lacks a reasonable basis, and the reasoning process is unreliable.\newline
\newline
Scoring Differences Explained:\newline
 - The main difference between 5 points and 4 points is whether the information extracted by the model fully supports the final result.\newline
 - The difference between 5 points and 4 points is whether the final conclusion is fully supported by the extracted information.\newline
 - The difference between 4 points and 3 points is whether a large part or a small part of the final conclusion is supported.\newline
 - The difference between 3 points and 2 points is whether at least one conclusion is supported. If so, it's 3 points; if not, but the information is related, it's 2 points.
 - The difference between 2 points and 1 point is whether the extracted and reasoned information is relevant to the final conclusion.\newline
\newline
V. Evaluation Examples\newline
\textbf{Example 1 (5 points, Lab Test)}\newline
1. Task Type:\newline
Predict the lab tests the patient should undergo.\newline
\newline
2. Model Reasoning Chain:\newline
\#\#Extraction:\newline
Source: Outpatient Record [2178-09-23 11:20:00]\newline
 - Key Information: Patient complains of fatigue and pale skin/mucous membranes.\newline
 - Analysis: Typical signs of anemia, requiring clarification of anemia type.\newline
Source: Lab Results [2178-09-22 15:45:00]\newline
 - Key Information: Hemoglobin 8.2 g/dL (normal 13-17), Creatinine 2.1 mg/dL (normal 0.7-1.3).\newline
 - Analysis: Moderate to severe anemia combined with abnormal kidney function suggests possible renal anemia.\newline
Source: Past Medical History [2177-05-12 09:00:00]\newline
 - Key Information: History of chronic kidney disease (CKD stage 3), not on dialysis.\newline
 - Analysis: Chronic kidney disease often leads to renal anemia. Erythropoietin levels need to be assessed.\newline
Source: Past Surgical History [2176-08-05 10:00:00]\newline
 - Key Information: Appendectomy, good recovery.\newline
 - Analysis: Not directly related to the current anemia evaluation. This is irrelevant information.\newline
\newline
\#\#Reasoning:\newline
Combining "anemia symptoms + significantly low hemoglobin + abnormal kidney function + chronic kidney disease," the following lab tests are recommended to clarify the anemia mechanism:\newline
Serum iron\newline
Ferritin\newline
Transferrin saturation\newline
Erythropoietin levels\newline
\newline
\#\#Final Result:\newline
Recommended Tests: Serum iron, Ferritin, Transferrin saturation, Erythropoietin levels.\newline
\newline
3. Score Explanation:\newline
Score: 5 points\newline
Reasoning:\newline
The first three pieces of extracted information ("fatigue + pale skin," "Hemoglobin 8.2 + Creatinine 2.1," "chronic kidney disease history") fully support the necessity of performing tests for iron metabolism and erythropoietin levels.\newline
The "past appendectomy" is extraneous information unrelated to the conclusion, but it does not affect the correctness of the conclusion.\newline
All key points of the conclusion are fully justified by the first three essential pieces of information, thus a score of 5 points.\newline
\newline
\textbf{Example 2 (4 points, Medication Prescription)}\newline
1. Task Type:\newline
Predict the patient's medication plan.\newline
\newline
2. Model Reasoning Chain:\newline
\#\#Extraction:\newline
Source: Inpatient Record [2180-02-15 08:10:00]\newline
 - Key Information: Type 2 diabetes mellitus, fasting blood glucose 12.5 mmol/L, postprandial blood glucose 16.0 mmol/L.\newline
 - Analysis: Blood glucose is significantly high; the current oral medication regimen is insufficient.\newline
Source: Lab Results [2180-02-14 07:50:00]\newline
 - Key Information: HbA1c 8.5\%, Creatinine 60 µmol/L (normal).\newline
 - Analysis: High HbA1c indicates the need for improved blood glucose control. Normal kidney function means GLP-1 agonists are a viable option.\newline
Source: Past Medication History [2179-10-10 09:00:00]\newline
 - Key Information: Taking oral Metformin 1000 mg twice daily and Gliclazide 80 mg once daily.\newline
 - Analysis: The current dual-drug therapy has failed to achieve the target, requiring an upgrade.\newline
\newline
\#\#Reasoning:\newline
Poor glycemic control (fasting 12.5, postprandial 16.0, HbA1c 8.5\%) → Needs escalation of antidiabetic treatment.\newline
Normal kidney function → GLP-1 agonists are a safe option.\newline
Has a history of hypertension and is taking Amlodipine → Needs to continue antihypertensive therapy.\newline
\newline
\#\#Final Result:\newline
Prescription: Add a GLP-1 receptor agonist, continue Amlodipine treatment.\newline
\newline
3. Score Explanation:\newline
Score: 4 points\newline
Reasoning:\newline
Support for adding GLP-1: The information "fasting 12.5 mmol/L, HbA1c 8.5\%" clearly suggests that the current oral medications are insufficient, and an escalation to a GLP-1 is reasonable. Additionally, "normal kidney function" proves this drug can be used safely. This part of the conclusion aligns perfectly with the extracted information.\newline
Support for continuing Amlodipine: The extracted information "history of hypertension and is taking Amlodipine" indicates the patient is already on antihypertensive treatment. This sub-conclusion is also directly supported by the extracted information.\newline
Overall: In this case, both parts of the conclusion—"add GLP-1" and "continue Amlodipine"—are supported by the corresponding information. The model's mention of "continuing Amlodipine" is a reasonable inference (most patients with T2DM have concurrent hypertension) but is not directly proven by the extracted information.\newline
Therefore, the majority of the conclusion (the GLP-1 part) is fully supported by the information, while a small part of the conclusion (continuing Amlodipine) is a reasonable but unsupported inference. This fits the criteria for a "4-point" scenario.\newline
\newline
\textbf{Example 3 (3 points, Diagnosis)}\newline
1. Task Type:\newline
Predict the patient's diagnosis.\newline
\newline
2. Model Reasoning Chain:\newline
\#\#Extraction:\newline
Source: Outpatient Record [2179-05-10 10:30:00]\newline
 - Key Information: Patient complains of sore throat, cough, fever, and fatigue for 3 days.\newline
 - Analysis: Suggests symptoms of an upper respiratory tract infection (URTI).\newline
Source: Physical Examination [2179-05-10 10:45:00]\newline
 - Key Information: Temperature 38.5°C, pharyngeal congestion, bilateral tonsillar enlargement (Grade II).\newline
 - Analysis: Supports acute tonsillitis or acute pharyngitis.\newline
Source: Lab Results [2179-05-10 11:00:00]\newline
 - Key Information: White blood cell count 12.5×10$^9$/L (normal 4-10), Neutrophils 80\% (normal 50-75\%).\newline
 - Analysis: Both white blood cells and neutrophils are elevated, suggesting a bacterial infection.\newline
\newline
\#\#Reasoning:\newline
Sore throat, fever, cough, temperature 38.5°C, pharyngeal congestion, tonsillar enlargement → Suggests URTI.\newline
Elevated white blood cells and neutrophils → Suggests bacterial infection.\newline
Combining the above information, the diagnosis is "acute upper respiratory tract infection with bacterial infection."\newline
Given the symptoms and signs, acute bronchitis is also a possibility.\newline
\newline
\#\#Final Result:\newline
Diagnosis: Acute upper respiratory tract infection with bacterial infection, acute bronchitis.\newline
\newline
3. Score Explanation:\newline
Score: 3 points\newline
Reasoning:\newline
The extracted information (sore throat, fever, elevated temperature, pharyngeal congestion, elevated white blood cells and neutrophils) fully supports the diagnosis of "acute upper respiratory tract infection with bacterial infection." This part of the conclusion is highly consistent with the information and is correct.\newline
The diagnostic conclusion also includes "acute bronchitis," but the extracted information does not contain direct evidence to support this diagnosis. For example, there is no mention of worsening cough, productive cough, or abnormal findings on lung auscultation (such as rales or rhonchi).\newline
While the majority of the conclusion ("acute upper respiratory tract infection with bacterial infection") is supported by the information, the diagnosis of "acute bronchitis" lacks a basis. According to the scoring criteria, "The extracted information only supports a small part of the conclusion, and the final result contains a large portion of content that is not adequately supported," therefore, the score is 3 points.\newline

\textbf{Example 4 (2 points, Medication Prescription)}\newline
1. Task Type:\newline
Predict the patient's medication plan.\newline
\newline
II. Model Reasoning Chain:\newline
\#\#Extraction:\newline
Source: Inpatient Record [2180-03-05 09:00:00]\newline
 - Key Information: Patient admitted due to palpitations and chest tightness. ECG shows atrial fibrillation.\newline
 - Analysis: The patient has a cardiac arrhythmia.\newline
Source: Past Medical History [2178-01-15 14:00:00]\newline
 - Key Information: History of hypertension, long-term use of Irbesartan.\newline
 - Analysis: Blood pressure control needs to be continued.\newline
Source: Lab Results [2180-03-05 10:30:00]\newline
 - Key Information: Normal kidney function, normal liver function.\newline
 - Analysis: Facilitates drug selection.\newline
\newline
\#\#Reasoning:\newline
The patient has atrial fibrillation and needs heart rate control.\newline
Long-term hypertension requires continued blood pressure management.\newline
Considering all factors, Metoprolol is recommended.\newline
\newline
\#\#Final Result:\newline
Prescription: Metoprolol.\newline
\newline
3. Score Explanation:\newline
Score: 2 points\newline
Reasoning:\newline
The extracted information—"palpitations, chest tightness, atrial fibrillation"—does indicate a cardiac arrhythmia, and Metoprolol is a commonly used first-line drug for treating atrial fibrillation. From a pharmacological perspective, this conclusion is related to the extracted information.\newline
However, the extracted information does not provide sufficient evidence to support Metoprolol as the sole or preferred drug. The treatment plan for atrial fibrillation requires a comprehensive consideration of the patient's heart rate control needs (type and rate of A-fib), stroke risk (CHA$_2$DS$_2$-VASc score), presence of comorbidities (like heart failure), and blood pressure levels. Based solely on "atrial fibrillation" and "hypertension," one cannot directly conclude that Metoprolol should be prescribed, as there are various other drug options (e.g., Diltiazem, Verapamil).\newline
The conclusion "Metoprolol" lacks sufficient evidence to be fully supported. It might be a plausible treatment, but it is not the only or best choice that the extracted information can definitively determine. This aligns with the scoring criterion: "The extracted information is related to the final result, but it fails to support the conclusion."\newline
\newline
\textbf{Example 5 (1 point, Lab Test)}\newline
1. Task Type:\newline
Predict the lab tests the patient should undergo.\newline
\newline
2. Model Reasoning Chain:\newline
\#\#Extraction:\newline
Source: Outpatient Record [2179-11-20 15:00:00]\newline
 - Key Information: Patient complains of abdominal pain and diarrhea.\newline
 - Analysis: Suggests a gastrointestinal issue.\newline
Source: Physical Examination [2179-11-20 15:20:00]\newline
 - Key Information: Abdominal palpation shows no significant tenderness, bowel sounds are active.\newline
 - Analysis: Suggests no organic lesion or inflammation in the abdomen.\newline
Source: Past Medication History [2179-11-19 18:00:00]\newline
 - Key Information: Recently took Aspirin.\newline
 - Analysis: No direct correlation with diarrhea.\newline
\newline
\#\#Reasoning:\newline
Patient with abdominal pain and diarrhea → Gastrointestinal symptoms.\newline
No tenderness on abdominal palpation → Rules out conditions like acute peritonitis.\newline
Patient took Aspirin → May affect the gastrointestinal tract.\newline
Based on the above information, an autoimmune disease is suspected.\newline
\newline
\#\#Final Result:\newline
Recommended Tests: Antinuclear antibody (ANA) profile, Complement, Rheumatoid factor.\newline
\newline
3. Score Explanation:\newline
Score: 1 point\newline
Reasoning:\newline
The extracted information is "abdominal pain, diarrhea," "no tenderness on abdominal palpation," and "took Aspirin." This information points to common gastrointestinal symptoms and potential drug-related GI discomfort.\newline
However, the final conclusion, "Recommended Tests: Antinuclear antibody profile, Complement, Rheumatoid factor," is used to diagnose autoimmune diseases (e.g., Systemic Lupus Erythematosus, Rheumatoid Arthritis).\newline
There are no symptoms or signs in the extracted information (e.g., joint pain, rash, oral ulcers) that are related to autoimmune diseases. Linking "abdominal pain and diarrhea" to tests for autoimmune diseases lacks any reasonable basis.\newline
The conclusion is severely irrelevant to the extracted information, and the reasoning process is highly unreliable. Therefore, the score is 1 point.
\end{prompt}

\begin{prompt}
\label{prompt: reasoning}
\textbf{Reasoning Chain Synthesis Prompt for GPT-4o} \newline
======================================== \newline
\# Patient EHR Context \# \newline
\{context\} \newline
 \newline
======================================== \newline
\# Retrieved Medical Knowledge \# \newline
\{medical\_knowledge\} \newline
 \newline
======================================== \newline
\# Ground Truth \# \newline
\{ground\_truth\} \newline
 \newline
======================================== \newline
\# Task \# \newline
\{task\} \newline
======================================== \newline
\newline
\# Data Description \newline
- \# Patient EHR Context \#: Contains all medical events and content from the patient's hospitalization journey. \newline
- \# Retrieved Medical Knowledge \#: Contains Patient EHR Context elements and their relationships potentially relevant to the Ground Truth entity. \newline
- \# Ground Truth \#: Contains the correct answers for this task; your predictions must exactly match these. \newline
- \# Task \#: Contains the specific task description; you need to complete the task based on the information in \# Patient EHR Context \#. \newline
\newline
\# Instructions \newline
Please provide a logically rigorous medical reasoning process so that the \# Ground Truth \# can be derived from the content in \# Patient EHR Context \# and \# Task \#.

\# Requirements \newline
The reasoning process should include three stages: Extraction, Reasoning, and Final Results. \newline
\newline
\#\# Extraction \newline
- In this stage, extract and identify each piece of "key information" from the \# Patient EHR Context \# according to the provided \# Retrieved Medical Knowledge \#. \newline
- Don't pay attention to information that you think is not helpful for the reasoning.  \newline
- Each step in the Extraction stage should follow the format below, **You need to specify the event name and time for each extracted information**: \newline
    **Event Name [Event Time]**: list the information extracted from the event and analyze the potential relationship between the key information and the ground truth. \newline
 \newline
\#\# Reasoning \newline
- Analyze the relationship between the context information and the item in \# Ground Truth \# in a very specific and professional manner, providing detailed reasoning steps. \newline
- Your analysis should include the item in \# Ground Truth \# as much as possible. Items that cannot be inferred from the context can be omitted. \newline
- Do not use the word "maybe", "possible" or "though" in the generated reasoning. You should do your best to find all the supporting information you can to ensure the correctness of your reasoning. \newline
- The reasoning process should be concise and rigorous, and each step should explain the specific medical knowledge involved, making the reasoning process more credible. \newline
- All reasoning must be based on the context information to infer the items in the ground truth and no reverse inference can be performed. \newline
\newline
\#\# Final Results
- Provide the final result for the task. **Note that the final result should only contain the items contained in \# Gound Truth \# that have been correctly inferred in. \newline
\#\# Reasoning stage.** \newline
- Each item in the \#\# Final Results should be contained in the \# Gound Truth \# with extactly same string. \newline
 \newline
\# Important Notes!!! \newline
- **For each piece of \# Retrieved Medical Knowledge \# that is relevant to completing the \# Task \#, locate the exact position of its first item within the \# Patient EHR Context \# and explicitly annotate it during the Extraction phase to ensure a more thorough analysis.** \newline
- **During the \#\# Reasoning stage, remember to analyze very carefully how each item in \# Ground Truth \# is inferred** \newline
- **Most importantly, integrate references to \# Ground Truth \# and \# Retrieved Medical Knowledge \# in an implicit manner. At any point in the reasoning process, do not use phrases such as “according to the medical knowledge above”, "as shown in ground truth" or any wording that reveals you are aware of the underlying medical knowledge or the ground truth.** \newline
 \newline
\# Output Format \newline
\#\# Extraction \newline
[YOUR OUTPUT] \newline
 \newline
\#\# Reasoning \newline
[YOUR OUTPUT] \newline
 \newline
\#\# Final Results \newline
[YOUR OUTPUT]
\end{prompt}

\begin{table}[!h]
    \centering
    \includegraphics[width=1\linewidth]{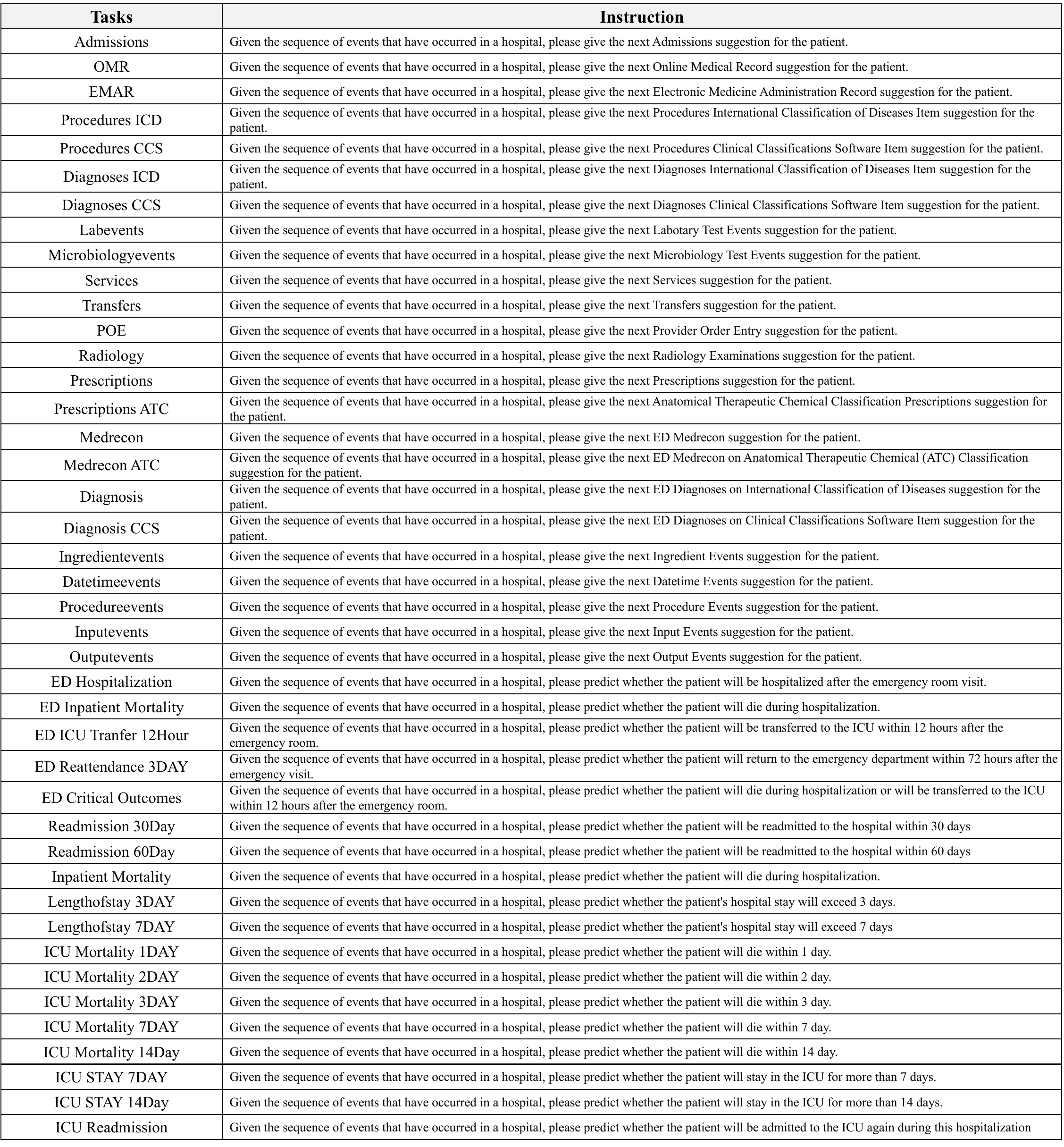}
    \vspace{1pt}
    \caption{\textbf{Instruction for each task in \benchname.}}
    \label{fig: instruction}
\end{table}

\begin{table}[!h]
    \centering
    \includegraphics[width=1\linewidth]{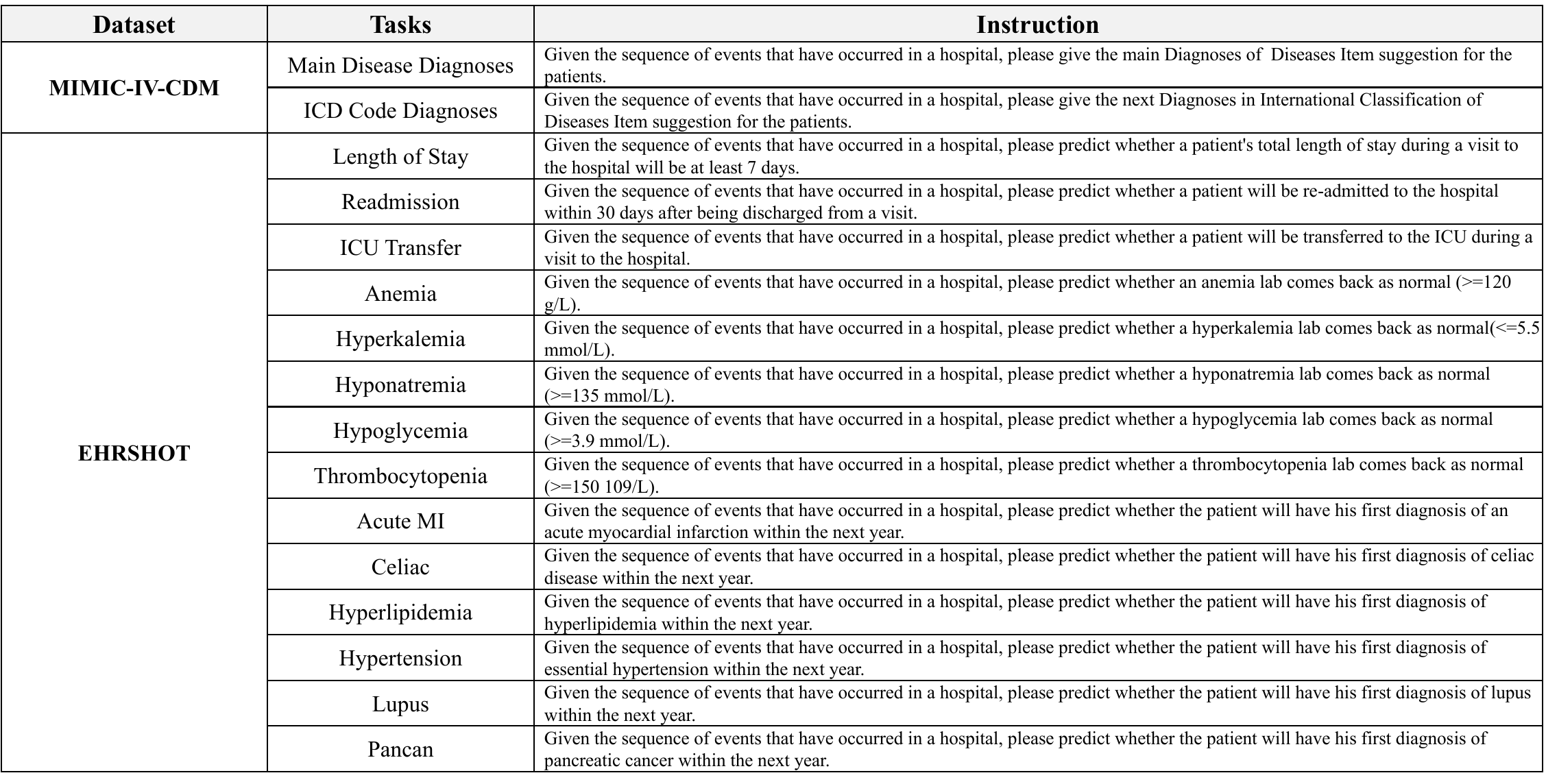}
    \vspace{1pt}
    \caption{\textbf{Instruction for each task in MIMIC-IV-CDM and EHRSHOT.}}
    \label{fig: ood_instruction}
\end{table}

\newpage
\subsection{Case Collection}
% ...

% ...
\begin{case}
\label{case: single item}
\textbf{Markdown Format for Event with Single Item} \newline
\#\# Event Name [Event Time] \newline
- Info Key 1: Info Value 1 \newline
- Info Key 2: Info Value 2 \newline
... \newline
- Info Key N: Info Value N
\end{case}

\begin{case}
\label{case: multiple item}
\textbf{Markdown Format for Event with Multiple Item} \newline
\#\# Event Name [Event Time]\newline
| Info Key 1 | Info Key 2 | ... | Info Key N |\newline
| ---------- | ---------- | --- | ---------- |\newline
| Info Value | Info Value | ... | Info Value |\newline
| Info Value | Info Value | ... | Info Value |\newline
...\newline
| Info Value | Info Value | ... | Info Value |
\end{case}

\begin{figure}[!t]
    \centering
    \includegraphics[width=1\linewidth]{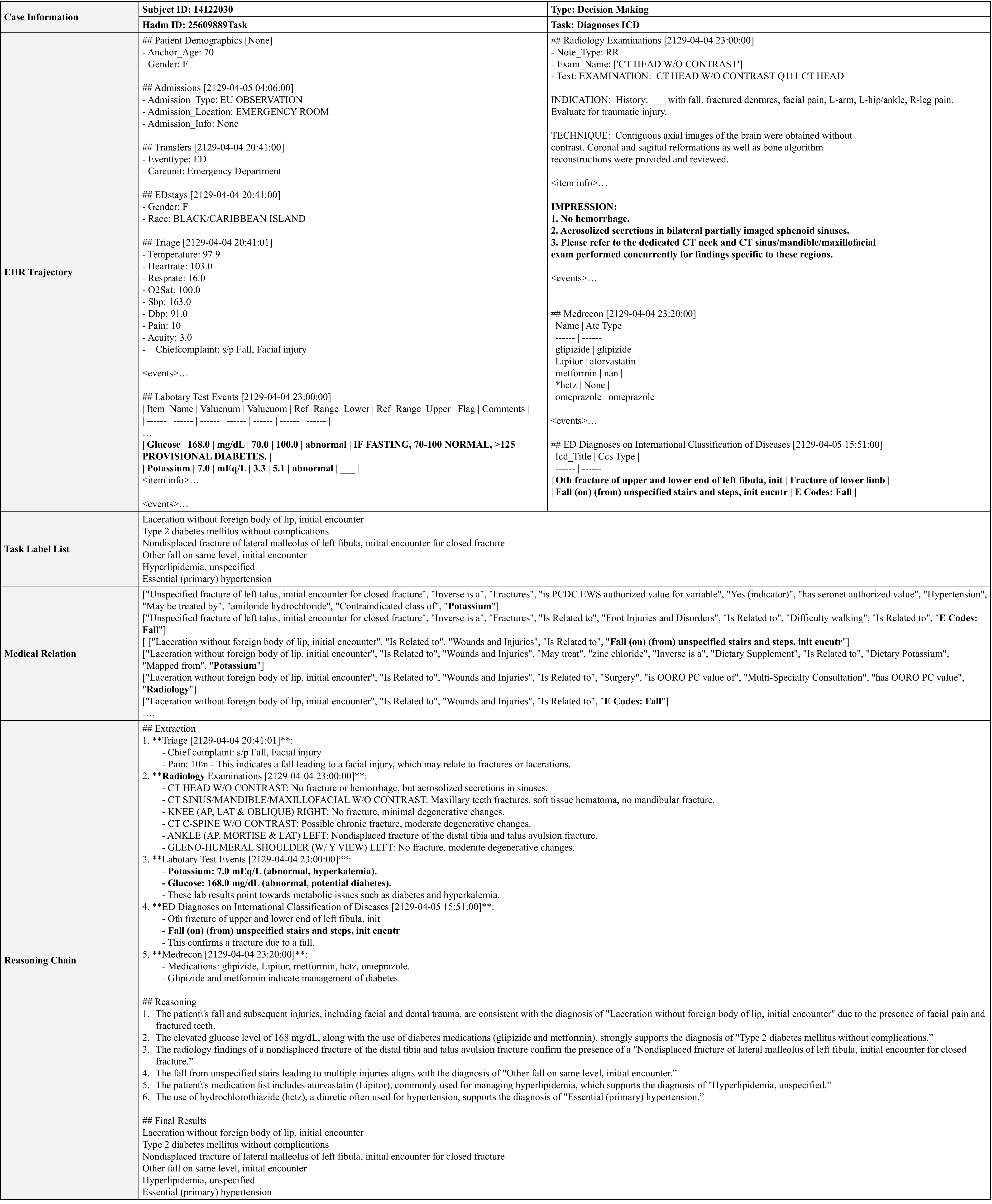}
    \caption{A case study of EHR Trajectory, Medical Relation, and Reasoning Chain. (a) EHR Trajectory for a patient, where <events>... and <item info>... represent the omission of a large amount of information for display purposes. (b) Medical Relation, showing the connections between the context medical entities and target items. (c) Reasoning Chain, detailing the process of inferring a diagnosis from the patient's EHR data. The parts highlighted in bold are the content commonly found in the EHR Trajectory, Medical Relation, and Reasoning Chain. This indicates that the medical graph is effective in identifying valid medical entities from the trajectory and using them to enhance reasoning.}
    \label{fig: case}
\end{figure}

\begin{case}
\label{case: ehrshot}
\textbf{Free-text Input Example of EHRSHOT}\newline
\#\# Person [1991-04-12 00:00:00]\newline
- Birth\newline
- White\newline
- Not Hispanic or Latino\newline
- MALE\newline
\newline
\#\# Measurement [2015-02-24 02:25:00]\newline
| Item\_Name | Valuenum | Valueuom | Ref\_Range\_Lower | Ref\_Range\_Upper | Flag | \newline
| ------ | ------ | ------ | ------ | ------ | ------ |\newline
| Lactate [Moles/volume] in Blood | 0.6000000238418579 | mmol/L | nan | nan | nan |\newline
| Lactate [Mass/volume] in Blood | 0.6000000238418579 | mmol/L | nan | nan | nan |\newline
| Glasgow coma scale | 15.0 | nan | nan | nan | nan |\newline
| Mean blood pressure | 103.0 | mmHg | nan | nan | nan |\newline
| Pain severity [Score] Visual analog score | 6.0 | nan | nan | nan | nan |\newline
| Respiratory rate | 18.0 | breaths/min | 12 | 18 | normal |\newline
| Systolic blood pressure | 152.0 | mmHg | 90 | 140 | abnormal |\newline
| Body temperature | 98.4000015258789 | F | 95 | 100.4 | normal |\newline
| Diastolic blood pressure | 79.0 | mmHg | 60 | 90 | normal |\newline
| Heart rate | 97.0 | bpm | 60 | 100 | normal |\newline
\newline
\#\# Observation [2015-02-24 02:25:00]\newline
| Item\_Name | Valuenum | Valueuom | Ref\_Range\_Lower | Ref\_Range\_Upper | Flag |\newline
| ------ | ------ | ------ | ------ | ------ | ------ |\newline
| Body temperature measurement site | 1.0 | nan | nan | nan | nan |\newline
| Oxygen saturation | 100.0 | \% | 95 | 100 | normal |\newline
\newline
\#\# Drug\_Exposure [2015-02-24 04:22:00]\newline
- oxycodone hydrochloride 1 MG/ML Oral Solution\newline
- ibuprofen 400 MG Oral Tablet\newline
- gabapentin 300 MG Oral Capsule\newline
- levetiracetam 500 MG Oral Tablet\newline
- 1 ML hydromorphone hydrochloride 2 MG/ML Prefilled Syringe\newline
- acetaminophen 325 MG Oral Tablet\newline
- 0.4 ML enoxaparin sodium 100 MG/ML Prefilled Syringe\newline
- zolpidem tartrate 5 MG Oral Tablet\newline
- triamcinolone acetonide 1 MG/ML Topical Cream\newline
- oxycodone hydrochloride 5 MG Oral Tablet\newline
\newline
\#\# Procedure\_Occurrence [2015-02-24 23:59:00]\newline
- Routine venipuncture\newline
- Taking patient vital signs assessment\newline
- Ambulating patient\newline
\newline
\#\# Condition\_Occurrence [2015-02-25 11:07:00]\newline
- Tobacco dependence syndrome\newline
- Benign neoplasm of colon\newline
- Inflammatory dermatosis\newline
- Epilepsy
\end{case}

\begin{case}
\label{case: mimic-iv-cdm}
\textbf{Free-text Input Example of MIMIC-IV-CDM}\newline
\#\# Patient Demographics\newline
- Patient History: \_\_\_ s/p emergency tissue AVR and type A aortic dissection repair \_\_\_ and sternal washout \_\_\_, c/b tamponade requiring  clot evacuation \_\_\_, trach \_\_\_, PEG \_\_\_. Overall, other  complications included seizures, renal failure requiring CRRT,  shock liver, HITT, sepsis, afib, left colonic ischemia. Discharged at end of \_\_\_ and returned to \_\_\_ in \_\_\_, at  which time he underwent a bronch.   Has been at rehab. Was on tube feeds until several weeks ago and  now only on regular diet. He has been eating, but decreased  appetite and intake due to post-prandial epigastric abdominal  pain. This has been occuring for approximately three weeks, and  he initially attributed it to indigestion. He had routine follow  up in thoracic clinic with Dr \_\_\_, at which time his PEG tube  was removed. Then he was then sent to the ED for his abdominal pain. PEG site was clean prior to removal.\newline
- Past Medical History: Hypertension Appendectomy  Right facial cyst drainage \newline   
- Social History: \_\_\_ Family History: None\newline
- Physical Examination: Physical exam  VS: 98.6, 80, 135/72, 18, 96\% RA Gen: NAD CV: RRR Pulm: CTA b/l Abd: soft, nondistended. slightly tender in epigastric area and right upper quadrant without rebound/guarding/rigidity Ext: left hand with ischemic digits, b/l \_\_\_ wrapped with gauze PE\newline
\newline

\#\# Labotary Test Events\newline
| Item\_name | Valuenum | Valueuom | Ref\_range\_lower | Ref\_range\_upper |\newline
| ------ | ------ | ------ | ------ | ------ |\newline
| Alanine Aminotransferase (ALT) | 18.0 | IU/L | 0.0 | 40.0 |\newline
| PTT | 32.8 | sec | 25.0 | 36.5 |\newline
| PT | 14.9 | sec | 9.4 | 12.5 |\newline
| INR(PT) | 1.4 | None | 0.9 | 1.1 |\newline
| White Blood Cells | 25.8 | K/uL | 4.0 | 11.0 |\newline
| Red Blood Cells | 3.84 | m/uL | 4.6 | 6.2 |\newline
| RDW | 14.1 | \% | 10.5 | 15.5 |\newline
| Platelet Count | 177.0 | K/uL | 150.0 | 440.0 |\newline
| Neutrophils | 91.5 | \% | 50.0 | 70.0 |\newline
| Monocytes | 3.6 | \% | 2.0 | 11.0 |\newline
| MCV | 92.0 | fL | 82.0 | 98.0 |\newline
| MCHC | 33.0 | \% | 31.0 | 35.0 |\newline
| Lymphocytes | 4.8 | \% | 18.0 | 42.0 |\newline
| Hemoglobin | 11.6 | g/dL | 14.0 | 18.0 |\newline
| Hematocrit | 35.2 | \% | 40.0 | 52.0 |\newline
| Eosinophils | 0.1 | \% | 0.0 | 4.0 |\newline
| MCH | 30.2 | pg | 27.0 | 32.0 |\newline
| Light Green Top Hold | HOLD. | None | None | None |\newline
| Albumin | 3.3 | g/dL | 3.5 | 5.2 |\newline
| Alkaline Phosphatase | 117.0 | IU/L | 40.0 | 130.0 |\newline
| Anion Gap | 15.0 | mEq/L | 8.0 | 20.0 |\newline
| Basophils | 0.1 | \% | 0.0 | 2.0 |\newline
| Bicarbonate | 28.0 | mEq/L | 22.0 | 32.0 |\newline
| Bilirubin, Total | 1.2 | mg/dL | 0.0 | 1.5 |\newline
| Chloride | 94.0 | mEq/L | 96.0 | 108.0 |\newline
| Asparate Aminotransferase (AST) | 20.0 | IU/L | 0.0 | 40.0 |\newline
| Estimated GFR (MDRD equation) | Using this patient's age, gender, and serum creatinine value of 1.2,.  Estimated GFR = 61 if non African-American (mL/min/1.73 m2).  Estimated GFR = 74 if African-American (mL/min/1.73 m2).  For comparison, mean GFR for age group 60-69 is 85 (mL/min/1.73 m2).  GFR<60 = Chronic Kidney Disease, GFR<15 = Kidney Failure. | None | None | None |\newline
| Glucose | 141.0 | mg/dL | 70.0 | 100.0 |\newline
| Lipase | 15.0 | IU/L | 0.0 | 60.0 |\newline
| Creatinine | 1.2 | mg/dL | 0.5 | 1.2 |\newline
| Potassium | 4.6 | mEq/L | 3.3 | 5.1 |\newline
| Sodium | 132.0 | mEq/L | 133.0 | 145.0 |\newline
| Urea Nitrogen | 27.0 | mg/dL | 6.0 | 20.0 |\newline
| Specimen Type | VEN. | None | None | None |\newline
| Lactate | 1.3 | mmol/L | 0.5 | 2.0 |\newline
| Protein | 30.0 | mg/dL | None | None |\newline
| Yeast | NONE | None | None | None |\newline
| WBC | <1. | None | 0.0 | 5.0 |\newline
| Urobilinogen | NEG. | None | 0.2 | 1.0 |\newline
| Urine Color | Yellow. | None | None | None |\newline
| Specific Gravity | >1.050*. | None | 1.001 | 1.035 |\newline
| RBC | 3.0 | \#/hpf | 0.0 | 2.0 |\newline
| pH | 6.0 | units | 5.0 | 8.0 |\newline
| Urine Appearance | Clear. | None | None | None |\newline
| Leukocytes | NEG. | None | None | None |\newline
| Ketone | NEG. | None | None | None |\newline
| Glucose | NEG. | None | None | None |\newline
| Epithelial Cells | 0.0 | \#/hpf | None | None |\newline
| Blood | TR. | None | None | None |\newline
| Bilirubin | NEG. | None | None | None |\newline
| Bacteria | NONE. | None | None | None |\newline
| Nitrite | NEG. | None | None | None |\newline
| Calcium, Total | 9.2 | mg/dL | 8.4 | 10.3 |\newline
| Magnesium | 1.8 | mg/dL | 1.6 | 2.6 |\newline
| Phosphate | 3.4 | mg/dL | 2.7 | 4.5 |\newline
\newline

\#\# Microbiology Test Events\newline
| Item\_name | Valuestr |\newline
| ------ | ------ |\newline
| Blood Culture, Routine | NO GROWTH. |\newline
| ANAEROBIC CULTURE | NO ANAEROBES ISOLATED. |\newline
| FLUID CULTURE | PSEUDOMONAS AERUGINOSA, ENTEROCOCCUS SP. |\newline
\newline

\#\# Radiology Examinations\newline
| Exam\_name | Text |\newline
| ------ | ------ |\newline
| LIVER OR GALLBLADDER US (SINGLE ORGAN) | EXAMINATION:
LIVER OR GALLBLADDER US (SINGLE ORGAN):

TECHNIQUE:
Grey scale and color Doppler ultrasound images of the abdomen were
obtained.

FINDINGS:

LIVER:
The hepatic parenchyma appears within normal limits. The contour of the
liver is smooth.  Again seen is a hyperechoic lesion measuring approximately 6
mm in segment V of the liver, unchanged since prior study and likely
represents a hemangioma. The additional hemangioma seen on previous study is
not clearly identified on today's exam. Main portal vein is patent with
hepatopetal flow. There is no ascites.

BILE DUCTS:
There is no intrahepatic biliary dilation. The CBD measures 6 mm.

GALLBLADDER:
The gallbladder is distended and filled with sludge.  There is
mild gallbladder wall edema measuring up to 6 mm. There is no pericholecystic
fluid.

PANCREAS:
The pancreas is not well seen secondary to overlying bowel gas.

KIDNEYS:
The right kidney measures 8.1 cm.  Survey views of the right kidney
do not demonstrate any masses, hydronephrosis, or stones.

RETROPERITONEUM:
Visualized portions of aorta and IVC are within normal
limits. |
\end{case}

\begin{case}
\label{case: mimic-bench}
\textbf{Free-text Input Example of \benchname}\newline
\# Patient Demographics [None]\newline
- Anchor\_Age: 88\newline
- Gender: F\newline
\newline
\#\# Admissions [2127-04-18 16:53:00]\newline
- Admission\_Type: EW EMER.\newline
- Admission\_Location: PROCEDURE SITE\newline
- Admission\_Info: None\newline
\newline
\#\# Provider Order Entry [2127-04-18 15:21:26]\newline
| Order\_Type | Order\_Subtype |\newline
| ------ | ------ |\newline
| Medications | nan |\newline
| ADT orders | Admit |\newline
| General Care | Vitals/Monitoring |\newline
| Nutrition | Diet Order |\newline
| General Care | Other |\newline
| Medications | nan |\newline
\newline
\#\# Services [2127-04-18 16:54:06]\newline
- Curr\_Service: CMED\newline
\newline
\#\# Transfers [2127-04-18 16:54:06]\newline
- Eventtype: admit\newline
- Careunit: Medicine/Cardiology\newline
\newline
\#\# Pharmacy [2127-04-18 16:57:15]\newline
| Medication | Proc\_Type | Status |\newline
| ------ | ------ | ------ |\newline
| Potassium Chloride | Unit Dose | Discontinued via patient discharge |\newline
| Potassium Chloride | Unit Dose | Discontinued via patient discharge |\newline
| Oxybutynin | Unit Dose | Discontinued via patient discharge |\newline
| Zolpidem Tartrate | Unit Dose | Discontinued via patient discharge |\newline
| Nitroglycerin SL | Unit Dose | Discontinued |\newline
| Sodium Chloride 0.9
| Hydrochlorothiazide | Unit Dose | Discontinued via patient discharge |\newline
| Potassium Chloride | Unit Dose | Discontinued via patient discharge |\newline
| Lisinopril | Unit Dose | Discontinued via patient discharge |\newline
| Multivitamins | Unit Dose | Discontinued via patient discharge |\newline
| Acetaminophen | Unit Dose | Discontinued via patient discharge |\newline
| Ferrous Sulfate | Unit Dose | Discontinued via patient discharge |\newline
| Aluminum-Magnesium Hydrox.-Simethicone | Unit Dose | Discontinued via patient discharge |\newline
| Pneumococcal Vac Polyvalent | Unit Dose | Discontinued |\newline
| Simvastatin | Unit Dose | Discontinued via patient discharge |\newline
| Aspirin | Unit Dose | Discontinued via patient discharge |\newline
| Diltiazem Extended-Release | Unit Dose | Discontinued via patient discharge |\newline
\newline
\#\# Prescriptions [2127-04-18 17:00:00]\newline
| Drug | Atc Type | Prod\_Strength | Dose\_Val\_Rx | Dose\_Unit\_Rx |\newline
| ------ | ------ | ------ | ------ | ------ |\newline
| Potassium Chloride | potassium chloride | 20mEq Packet | 20 | mEq |\newline
| Potassium Chloride | potassium chloride | 20mEq Packet | 40 | mEq |\newline
| Zolpidem Tartrate | zolpidem | 5mg Tablet | 5 | mg |\newline
| Nitroglycerin SL | glyceryl trinitrate | 0.3mg SL Tablet Bottle | 0.3 | mg |\newline
| Sodium Chloride 0.9
| Potassium Chloride | potassium chloride | 20mEq Packet | 60 | mEq |\newline
| Acetaminophen | paracetamol | 325mg Tablet | 650 | mg |\newline
| Aluminum-Magnesium Hydrox.-Simethicone | aluminium hydroxide | 30 mL UDCup | 30 | mL |\newline
| Pneumococcal Vac Polyvalent | nan | 25mcg/0.5mL Vial | 0.5 | mL |\newline
\newline
\#\# Provider Order Entry [2127-04-18 17:13:19]\newline
| Order\_Type | Order\_Subtype |\newline
| ------ | ------ |\newline
| Medications | nan |\newline
| Medications | nan |\newline
| Medications | nan |\newline
\newline
\#\# Pharmacy [2127-04-18 17:26:01]\newline
| Medication | Proc\_Type | Status |\newline
| ------ | ------ | ------ |\newline
| Fentanyl Citrate | Unit Dose | Expired |\newline
| Ibuprofen | Unit Dose | Expired |\newline
\newline
\#\# Provider Order Entry [2127-04-18 17:55:22]\newline
| Order\_Type | Order\_Subtype |\newline
| ------ | ------ |\newline
| Radiology | Ultrasound |\newline
| General Care | Other |\newline
| General Care | Other |\newline
| General Care | Other |\newline
| General Care | Other |\newline
| Blood Bank | Blood tests |\newline
| Lab | nan |\newline
| Lab | nan |\newline
| Radiology | General Xray |\newline
| Cardiology | ECG |\newline
| General Care | Other |\newline
| General Care | Other |\newline
\newline
\#\# Prescriptions [2127-04-18 18:00:00]\newline
| Drug | Atc Type | Prod\_Strength | Dose\_Val\_Rx | Dose\_Unit\_Rx |\newline
| ------ | ------ | ------ | ------ | ------ |\newline
| Fentanyl Citrate | fentanyl | 100mcg/2mL Amp | 25 | mcg |\newline
| Ibuprofen | ibuprofen | 600mg Tablet | 600 | mg |\newline
\newline
\#\# Provider Order Entry [2127-04-18 18:12:48]\newline
| Order\_Type | Order\_Subtype |\newline
| ------ | ------ |\newline
| General Care | Vitals/Monitoring |\newline
| General Care | Other |\newline
| General Care | Other |\newline
| General Care | Other |\newline
| General Care | Other |\newline
| General Care | Activity |\newline
| General Care | Other |\newline
| General Care | Other |\newline
| IV therapy | IV fluids |\newline
| Medications | nan |\newline
| General Care | Other |\newline
\newline
\#\# Pharmacy [2127-04-18 18:44:04]\newline
| Medication | Proc\_Type | Status |\newline
| ------ | ------ | ------ |\newline
| nan | IV Large Volume | Expired |\newline
| Atropine Sulfate | Unit Dose | Discontinued via patient discharge |\newline
\newline
\#\# Prescriptions [2127-04-18 19:00:00]\newline
| Drug | Atc Type | Prod\_Strength | Dose\_Val\_Rx | Dose\_Unit\_Rx |\newline
| ------ | ------ | ------ | ------ | ------ |\newline
| 5
| Sodium Bicarbonate | sodium bicarbonate | 50mEq Vial | 150 | mEq |\newline
| Atropine Sulfate | atropine | 1mg/10mL Syinge | 0.5 | mg |\newline
\newline
\#\# Provider Order Entry [2127-04-18 19:11:50]\newline
| Order\_Type | Order\_Subtype |\newline
| ------ | ------ |\newline
| Lab | nan |\newline
| Medications | nan |\newline
\newline
\#\# Pharmacy [2127-04-18 19:14:10]\newline
- Medication: Acetylcysteine 20
- Proc\_Type: Unit Dose\newline
- Status: Discontinued via patient discharge\newline
\newline
\#\# Labotary Test Events [2127-04-18 19:19:00]\newline
| Item\_Name | Valuenum | Valueuom | Ref\_Range\_Lower | Ref\_Range\_Upper | Flag | Comments |\newline
| ------ | ------ | ------ | ------ | ------ | ------ | ------ |\newline
| 
| INR(PT) | 1.2 | nan | 0.9 | 1.1 | abnormal | nan |\newline
| PT | 13.7 | sec | 10.4 | 13.4 | abnormal | nan |\newline
| PTT | 25.6 | sec | 22.0 | 35.0 | nan | nan |\newline
| Alanine Aminotransferase (ALT) | 9.0 | IU/L | 0.0 | 40.0 | nan | nan |\newline
| Albumin | 3.8 | g/dL | 3.4 | 4.8 | nan | nan |\newline
| Alkaline Phosphatase | 61.0 | IU/L | 39.0 | 117.0 | nan | nan |\newline
| Anion Gap | 13.0 | mEq/L | 8.0 | 20.0 | nan | nan |\newline
| Asparate Aminotransferase (AST) | 13.0 | IU/L | 0.0 | 40.0 | nan | nan |\newline
| Bicarbonate | 27.0 | mEq/L | 22.0 | 32.0 | nan | nan |\newline
| Bilirubin, Total | 0.2 | mg/dL | 0.0 | 1.5 | nan | nan |\newline
| Chloride | 99.0 | mEq/L | 96.0 | 108.0 | nan | nan |\newline
| Creatinine | 1.3 | mg/dL | 0.4 | 1.1 | abnormal | nan |\newline
| Estimated GFR (MDRD equation) | nan | nan | nan | nan | nan | Using this patient's age, gender, and serum creatinine value of 1.3,.  Estimated GFR = 39 if non African-American (mL/min/1.73 m2).  Estimated GFR = 47 if African-American (mL/min/1.73 m2).  For comparison, mean GFR for age group 70+ is 75 (mL/min/1.73 m2).  GFR<60 = Chronic Kidney Disease, GFR<15 = Kidney Failure. |\newline
| Glucose | 176.0 | mg/dL | 70.0 | 105.0 | abnormal | nan |\newline
| Lactate Dehydrogenase (LD) | 134.0 | IU/L | 94.0 | 250.0 | nan | nan |\newline
| Potassium | 3.8 | mEq/L | 3.3 | 5.1 | nan | nan |\newline
| Sodium | 135.0 | mEq/L | 133.0 | 145.0 | nan | nan |\newline
| Urea Nitrogen | 28.0 | mg/dL | 6.0 | 20.0 | abnormal | nan |\newline
| Hematocrit | 28.5 | 
| Hemoglobin | 9.6 | g/dL | 12.0 | 16.0 | abnormal | nan |\newline
| MCH | 29.2 | pg | 27.0 | 32.0 | nan | nan |\newline
| MCHC | 33.7 | 
| MCV | 87.0 | fL | 82.0 | 98.0 | nan | nan |\newline
| Platelet Count | 260.0 | K/uL | 150.0 | 440.0 | nan | nan |\newline
| RDW | 14.7 | 
| Red Blood Cells | 3.29 | m/uL | 4.2 | 5.4 | abnormal | nan |\newline
| White Blood Cells | 4.6 | K/uL | 4.0 | 11.0 | nan | nan |\newline
\newline
\#\# Prescriptions [2127-04-18 20:00:00]\newline
- Drug: Acetylcysteine 20
- Atc Type: acetylcysteine\newline
- Prod\_Strength: 800mg/4ml Vial\newline
- Dose\_Val\_Rx: 600\newline
- Dose\_Unit\_Rx: mg\newline
\newline
\#\# Labotary Test Events [2127-04-18 21:20:00]\newline
| Item\_Name | Valuenum | Valueuom | Ref\_Range\_Lower | Ref\_Range\_Upper | Flag | Comments |\newline
| ------ | ------ | ------ | ------ | ------ | ------ | ------ |\newline
| Bacteria | nan | nan | nan | nan | nan | NONE. |\newline
| Bilirubin | nan | mg/dL | nan | nan | nan | NEG. |\newline
| Blood | nan | nan | nan | nan | nan | MOD. |\newline
| Epithelial Cells | 1.0 | \#/hpf | nan | nan | nan | nan |\newline
| Glucose | nan | mg/dL | nan | nan | nan | NEG. |\newline
| Ketone | nan | mg/dL | nan | nan | nan | NEG. |\newline
| Leukocytes | nan | nan | nan | nan | nan | NEG. |\newline
| Nitrite | nan | nan | nan | nan | nan | NEG. |\newline
| pH | 7.5 | units | 5.0 | 8.0 | nan | nan |\newline
| Protein | nan | mg/dL | nan | nan | nan | NEG. |\newline
| RBC | nan | \#/hpf | 0.0 | 2.0 | nan | <1. |\newline
| Specific Gravity | 1.015 |   | 1.001 | 1.035 | nan | nan |\newline
| Transitional Epithelial Cells | nan | \#/hpf | nan | nan | nan | <1. |\newline
| Urine Appearance | nan | nan | nan | nan | nan | Hazy. |\newline
| Urine Color | nan | nan | nan | nan | nan | Straw. |\newline
| Urobilinogen | nan | mg/dL | 0.2 | 1.0 | nan | NEG. |\newline
| WBC | 3.0 | \#/hpf | 0.0 | 5.0 | nan | nan |\newline
| Yeast | nan | nan | nan | nan | nan | nan |\newline
\newline
\#\# Provider Order Entry [2127-04-19 02:55:46]\newline
- Order\_Type: Medications\newline
- Order\_Subtype: nan\newline
\newline
\#\# Pharmacy [2127-04-19 02:58:06]\newline
- Medication: Zolpidem Tartrate\newline
- Proc\_Type: Unit Dose\newline
- Status: Discontinued via patient discharge\newline
\newline
\#\# Prescriptions [2127-04-19 03:00:00]\newline
- Drug: Zolpidem Tartrate\newline
- Atc Type: zolpidem\newline
- Prod\_Strength: 5mg Tablet\newline
- Dose\_Val\_Rx: 5\newline
- Dose\_Unit\_Rx: mg\newline
\newline
\#\# Provider Order Entry [2127-04-19 05:41:43]\newline
- Order\_Type: Lab\newline
- Order\_Subtype: nan\newline
\newline
\#\# Labotary Test Events [2127-04-19 07:16:00]\newline
| Item\_Name | Valuenum | Valueuom | Ref\_Range\_Lower | Ref\_Range\_Upper | Flag | Comments |\newline
| ------ | ------ | ------ | ------ | ------ | ------ | ------ |\newline
| Hematocrit | 29.0 | 
| Creatinine | 1.4 | mg/dL | 0.4 | 1.1 | abnormal | nan |\newline
| Urea Nitrogen | 28.0 | mg/dL | 6.0 | 20.0 | abnormal | nan |\newline
\newline
\#\# Radiology Examinations [2127-04-19 09:49:00]\newline
- Note\_Type: RR\newline
- Exam\_Name: ['CHEST (PRE-OP PA \& LAT)']\newline
- Text: REASON FOR EXAMINATION:  Preoperative evaluation of the patient with aortic stenosis before aortic valve replacement. PA and lateral chest radiograph was compared to \_\_\_. Heart size is normal.  Mediastinal position, contour and width are stable except for dextroscoliosis, mild to moderate.  The lungs are clear except for linear bibasilar opacities, unchanged since \_\_\_, consistent with scarring.  There is no pleural effusion or pneumothorax.  The lateral view demonstrates contrast material in expected location of the distal esophagus that might be related to recent study involving the administration of oral contrast, please correlate with clinical history. There is no pleural effusion or pneumothorax demonstrated.\newline
\newline
\#\# Provider Order Entry [2127-04-19 12:11:28]\newline
| Order\_Type | Order\_Subtype |\newline
| ------ | ------ |\newline
| ADT orders | Discharge |\newline
| ADT orders | Discharge |\newline
\newline
\#\# Transfers [2127-04-19 13:10:00]\newline
- Eventtype: discharge\newline
- Careunit: nan
\end{case}

\end{document}